%% file: main.tex
\documentclass[sigconf,nonacm]{acmart}
\usepackage{times} 
\usepackage{helvet}  
\usepackage{courier}  
\usepackage{url} 
\usepackage{amsmath}
\usepackage{graphicx} 
\usepackage{amsthm}
\newtheorem{theorem}{Theorem}
\usepackage{subfigure}
\DeclareCaptionStyle{ruled}{labelfont=normalfont,labelsep=colon,strut=off}
\usepackage{algorithm}
\usepackage[noend]{algorithmic}
\usepackage{newfloat}
\usepackage{listings}
\usepackage{mdframed}
\floatstyle{ruled}
\newfloat{listing}{tb}{lst}{}
\floatname{listing}{Listing}

\usepackage{amsmath}
\usepackage{stmaryrd}
\usepackage{xspace}
\usepackage{booktabs}

\newcommand{\scenic}{{\sc Scenic}\xspace}
\newcommand{\semantics}[1]{\llbracket #1 \rrbracket}

\usepackage{todonotes}
\newcommand{\sebastian}[1]{\todo{!!}}

\title{Querying Labelled Data with Scenario Programs\\for Sim-to-Real Validation}
\author{Edward Kim, Jay Shenoy, Sebastian Junges,\\ Daniel Fremont$^{\dagger}$, Alberto Sangiovanni-Vincentelli, Sanjit A. Seshia \\
University of California, Berkeley\\
$^{\dagger}$ University of California, Santa Cruz}

\begin{document}

\begin{abstract}
Simulation-based testing of autonomous vehicles (AVs) has become an essential complement to road testing to ensure safety. Consequently, substantial research has focused on searching for failure scenarios in simulation. However, a fundamental question remains: are AV failure scenarios identified in simulation \textit{meaningful} in reality --- i.e., are they reproducible on the real system? Due to the sim-to-real gap arising from discrepancies between simulated and real sensor data, a failure scenario identified in simulation can be either a spurious artifact of the synthetic sensor data or an actual failure that persists with real sensor data. 
An approach to validate simulated failure scenarios is to identify instances of the scenario in a corpus of real data, and check if the failure persists on the real data.
To this end, we propose a formal definition of what it means for a labelled data item to match an abstract scenario, encoded as a scenario program using the \scenic{} probabilistic programming language.
Using this definition, we develop a querying algorithm which, given a scenario program and a labelled dataset, finds the subset of data matching the scenario.
Experiments demonstrate that our algorithm is accurate and efficient on a variety of realistic traffic scenarios, and scales to a reasonable number of agents.
\end{abstract}

\maketitle
\pagestyle{empty}

\input{sections/introduction}
\input{sections/overview}
\input{sections/background}

\input{sections/problem}

\input{sections/methodology}

\input{sections/experiment}
\input{sections/related}
\input{sections/conclusion}

{\small
\bibliographystyle{IEEEtranN}
\bibliography{reference}
}
\clearpage
\appendix
\input{sections/appendix}

\end{document}

%% file: sections/introduction.tex
\section{Introduction}\label{sec:intro}
Simulation-based testing is becoming a core element of assessing the safety of autonomous vehicles (AVs) by government and industry. For example, the National Highway Traffic Safety Administration stated that self-driving technology should be tested in simulation before deployment~\cite{nhtsa}, and Waymo  recently used simulation to support the claim that self-driving cars are safer than human drivers~\cite{waymo}. 
%Exploiting the ability to do risk-free and inexpensive testing in simulation, a large body of literature on simulators and simulation-based testing for AVs has been developed.
A number of open-source simulation environments designed to support automated AV testing are available~\cite{carla,lgsvl,airsim}, as well as simulators which focus on realistic rendering of specific types of sensors such as LiDAR and radar~\cite{lidarsim,radarsim}.
%where the realism is demonstrated by perception models trained on real sensor data being able to detect objects in the simulated sensor data.  
There are also a variety of black-box and white-box techniques to \textit{search} for failure scenarios causing an AV to violate its safety specifications~(e.g. \cite{sim-test1,sim-test2,sim-test3,sim-test4,verifai,kim-cvpr20}).

Despite these advances, the fundamental question: ``Are AV failures found in simulation \textit{meaningful}, i.e., are they reproducible on the real system?'' has not been answered. Because of the so-called ``sim-to-real gap'' --- the discrepancy between simulated and real sensor data and physics --- a failure scenario found in simulation can be spurious, with no similar failure being possible for the real system operating on real data. There is a need for techniques to \emph{validate} a simulated scenario against real sensor data. For brevity, we will refer to this problem as the \textit{scenario validation} problem.

Previous work on bridging the sim-to-real gap have focused on \textit{training}~\cite{sim-to-real1,sim-to-real2,sim-to-real3,sim-to-real4} AV components, such as perception and behavior prediction, only on simulated data and demonstrating that they perform robustly on real data. However, there is less work on \textit{testing}, i.e., to investigate how simulated AV behaviors transfer to real roads \textit{without} having to physically reconstruct these scenarios, an activity that is labor-intensive and not scalable. In contrast, we propose a \textit{data-driven} approach for the scenario validation problem.

To tackle this problem, we leverage two technological trends in the AV domain. First, both in academia and industry, there are active community efforts to collect and share large amounts of real sensor data with high quality \textit{labels}~\cite{bdd100k,kitti,apolloscape,waymodataset}.
Second, AV companies are \textit{mapping} road infrastructure, gathering detailed geometric information about intersections, lanes, sidewalks, etc.

Several such efforts have produced open-source datasets where real sensor data is synchronized with map information, such as nuScenes~\cite{nuscenes} and Argoverse~\cite{argoverse}.
Given these trends, in this paper, we assume to have \textit{access to a large set of labelled real sensor data and the corresponding map information}.

\begin{figure}
    \centering
    \includegraphics[width=\linewidth]{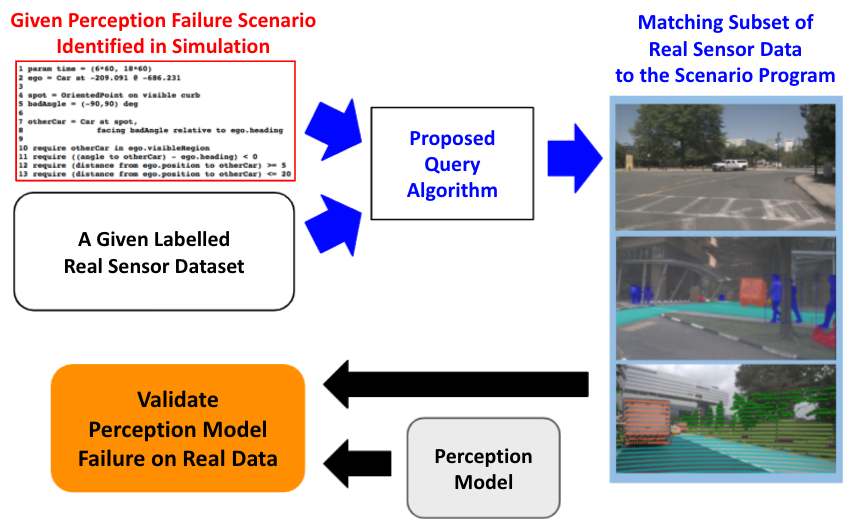}
    \caption{Using our proposed query algorithm to validate whether a failure of an AV's perception system previously identified in simulation persists in reality.}\label{fig:vision}
\end{figure}

Based on this assumption, we propose to validate a candidate failure scenario by \emph{querying} a dataset to find real-world instances of the scenario.
Specifically, we want a query algorithm that, given (1) a formal description of a failure scenario previously identified in simulation and (2) a set of labelled real sensor data with map information, outputs the subset of labels (and, therefore, of real sensor data) that \textit{matches} the scenario description. Developers could then test whether the AV failure is reproducible on this subset of real data to validate the failure scenario, as illustrated in Fig.~\ref{fig:vision}.

\begin{figure*}
  \centering
  \includegraphics[width=0.8\linewidth]{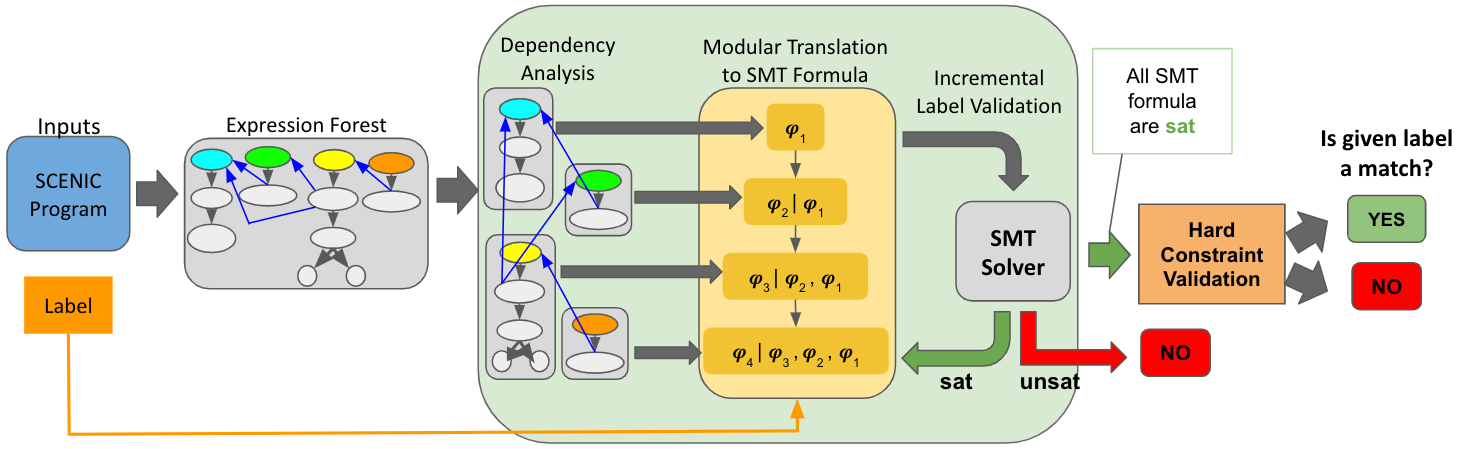}
  \caption{An overview of our algorithm to determine if a label matches a \scenic{} program.}
  \label{fig:overview}
\end{figure*}

\begin{figure}[h]
    \centering
    \includegraphics[width=\linewidth]{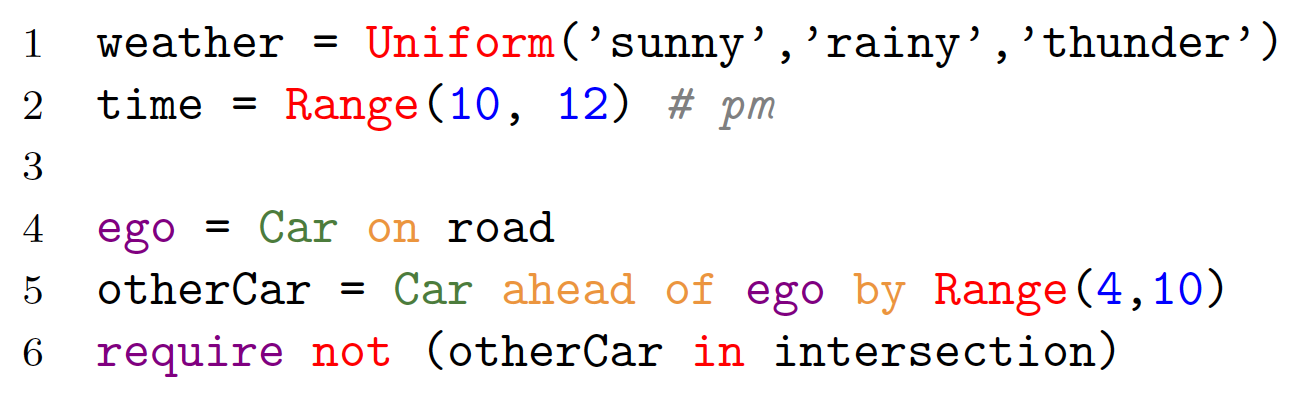}
    \caption{A \scenic{} program describing a car ahead of the ego car by 4--10 meters and which is not in an intersection.}
    \label{fig:scenic_program}
\end{figure}

In order to make the query problem precise and provide correctness guarantee, we must formally define what it means for a label to ``match'' a scenario.
To do this, we formally model scenarios as programs in the \scenic{} probabilistic programming language~\cite{scenic}, which is designed for scenario specification.
In this paper, we deal with \emph{static} scenarios describing the environment of the system at one moment in time, which is sufficient for testing many AV perception systems.
We can then define a label as matching a scenario if it has nonzero probability under the distribution defined by the \scenic{} program.
We develop an algorithm that checks this condition by encoding it as a series of \textit{Satisfiability Modulo Theories} (SMT)~\cite{barrett-smtbookch09} problems.
Because our algorithm queries labels, it is applicable to data of any type, including RGB, LiDAR, and radar (for simplicity we refer to ``images'' henceforth). Consequently, it can be used to validate failure scenarios for a wide variety of AV perception \emph{static} tasks (e.g. classification, detection, segmentation) utilizing various sensor types.

\paragraph{Contributions.}
In this paper, we provide:
\begin{itemize}
%   \item A novel problem formulation for querying a dataset with a scenario program, enabling simulated failure scenarios to be validated against real-world counterparts;
  \item A novel formulation of the problem of querying a labelled dataset against a formal scenario description, enabling simulated failure scenarios for a perception model to be validated against real-world counterparts;
  \item A sound SMT-based algorithm solving the query problem for static scenarios given by \scenic{} programs, which can be used to validate scenarios across a wide variety of perception tasks and sensor types;
  \item Experiments demonstrating the accuracy and scalability of our algorithm on realistic AV scenarios.
\end{itemize}

We start with an overview of our approach in Sec.~\ref{sec:overview}.
Section~\ref{sec:background} presents background allowing us to formally define the query problem in Sec.~\ref{sec:problem}.
We describe our algorithm in Sec.~\ref{sec:method} and our experiments in Sec.~\ref{sec:experiment}.
Finally, we discuss related work in Sec.~\ref{sec:relatedwork} and overall conclusions in Sec.~\ref{sec:concl}.

%% file: sections/overview.tex
% \begin{figure*}
%   \centering
%   \includegraphics[width=0.8\linewidth]{main_figures/overview.png}
%   \caption{An overview of our algorithm to determine if a label matches a \scenic{} program.}
%   \label{fig:overview}
% \end{figure*}

% \begin{figure}[h]
%     \centering
%     \includegraphics[width=\linewidth]{main_figures/example_scenic_program1.png}
%     \caption{A \scenic{} program describing a car ahead of the ego car by 4--10 meters and which is not in an intersection.}
%     \label{fig:scenic_program}
% \end{figure}

\section{Overview}\label{sec:overview}

%% Setting
Suppose we wish to query a label against the simple \scenic{} program in Fig.~\ref{fig:scenic_program}.
The program describes, for some fixed map with information about roads and intersections, a scenario where there is a car ahead of the ego vehicle but not in an intersection.
A label consists of semantic features such as time of day, weather conditions, and positions and orientations of vehicles.
%(note that interpreting this program requires map information such as the positions of roads and intersections).
%We assume that we are given a \emph{labeled} real-world data set.
For the particular example, assume that our (simplified) label $l$ consists of features $l_{e_x}, l_{e_y}, l_{e_h}, l_{e_{cl}}$ denoting the $xy$-coordinates, heading, and object class (e.g.~car, pedestrian) of the ego car respectively, and, likewise, features $l_{c_x}, l_{c_y}, l_{c_h}, l_{c_{cl}}$ for the other car.

%% Question
Whenever we query a label against a program, we want to answer the question: \emph{is the situation specified by the label an instance of the scenario in Fig.~\ref{fig:scenic_program}?}
More precisely, we ask whether the label can be obtained by instantiating the random variables in the scenario (i.e. for some choice of the ego's position ``on road'' and a distance between 4--10 meters).
% Our query is thus defined by the \scenic{} program. Our proposed approach is visualized in Fig.~\ref{fig:overview}. For the remainder of this section, we will refer to this figure along with the example \scenic{} program in Fig.~\ref{fig:scenic_program}.

%% Approach
Our approach to this problem is summarized in Fig.~\ref{fig:overview}. Before we go into details, let us clarify that in a nutshell, the approach is to translate the \scenic{} program and the label into constraints represented as a Satisfiability Modulo Theory (SMT) formula (see Sec.~\ref{sec:smt} for an overview of SMT).
The resulting formula will be satisfiable if and only if the given label matches the program. While the problem can be captured by a single (monolithic) SMT formula, the size of this formula quickly explodes for more complex scenarios. This blowup then prevents state-of-the-art methods from actually solving the formula.
To alleviate the scalability problem, we take advantage of the structure of the \scenic{} program to decompose it into several SMT formulas, determining incrementally whether parts of the label match the program. 
Below, we discuss Fig.~\ref{fig:overview} in slightly more detail.
%This approach improves scalability, avoiding the blow-up in constraints that would result from a direct translation of a complex scenario (e.g. one with many agents) into a monolithic SMT formula.

\begin{figure}
    \centering
    \includegraphics[width=\linewidth]{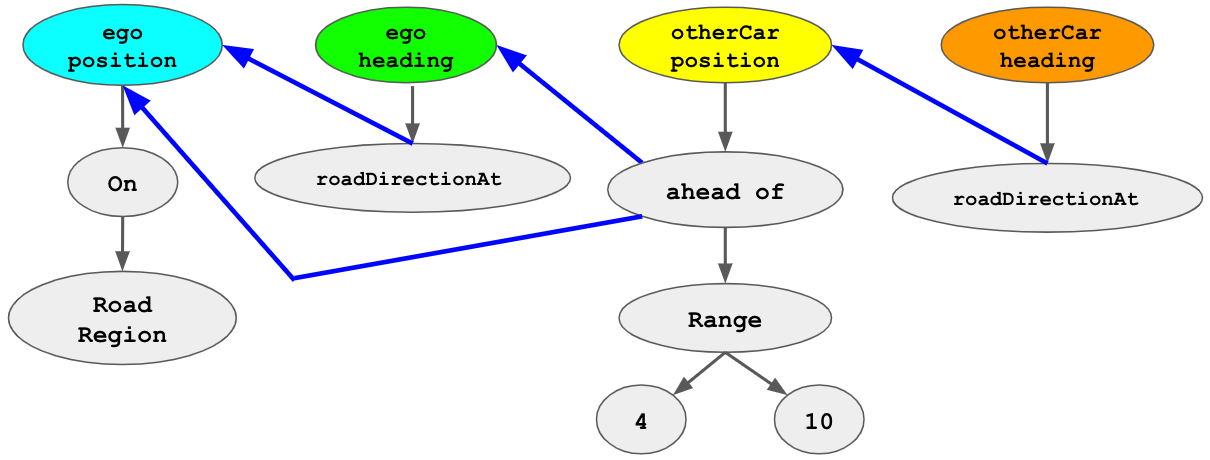}
    \caption{A partial expression forest for the \scenic{} program in Fig.~\ref{fig:scenic_program}. The 4 top nodes represent the semantic features, and the blue arrows show dependencies among them.}
    \label{fig:expression_tree}
\end{figure}

\paragraph{Expression forest}
To enable the decomposition process, the first step is operate on the internal representation of a \scenic{} program.
More precisely, the \scenic{} compiler converts a program into an \emph{expression forest}: a simplified forest for the program in Fig.~\ref{fig:scenic_program} is shown in Fig.~\ref{fig:expression_tree}.
The expression forest is made up of a set of expression trees, each of which essentially corresponds to the syntax used to define the distribution of one of the semantic features.
For example, the leftmost tree in Fig.~\ref{fig:expression_tree} shows that the ego's position is defined to be a uniformly random point in the \texttt{road} region of the map.

\paragraph{Dependency analysis}
Expression trees can have dependency relations with each other, as shown by the bold blue arrows in Fig.~\ref{fig:expression_tree}.
Such dependencies naturally occur in scenario modeling. For example, to compute the \texttt{otherCar}'s position (which is \texttt{ahead of} the ego car), we need to first know the ego's position and heading.
Our dependency analysis uses the \scenic{} expression forest to sort the list of semantic features in dependency order: in Fig.~\ref{fig:expression_tree}, the ego's position is first since it depends on no other features, while the \texttt{otherCar}'s heading is last.

\paragraph{Modular translation}
Next, we \textit{modularly} translate each expression tree and \textit{incrementally} check its consistency with the label in dependency order.
Here, we start with the ego's position, symbolically representing its coordinates by variables $e_x$ and $e_y$.
We encode its definition given by the leftmost expression tree in Fig.~\ref{fig:expression_tree} with the constraint $\texttt{On}(\texttt{RoadRegion},e_x,e_y)$, where \texttt{On} is a predicate requiring that the point $(e_x, e_y)$ is contained in the polygon that defines the \texttt{RoadRegion}.
We can then check whether the value of this feature given in the label is in fact possible in the \scenic{} program by checking the satisfiability of the SMT formula
 \[\varphi_{1}=\texttt{On}(\texttt{RoadRegion},e_x,e_y)\land(e_x=l_{e_x})\land (e_y=l_{e_y})\]
where as above $l_{e_x}$ and $l_{e_y}$ are the label's concrete values for $e_x$ and $e_y$, and $\land$ represents a logical AND.

\paragraph{Incremental validation}
If the formula is unsatisfiable, the observed label cannot be generated by the \scenic{} program, and so does not match the scenario.
If the formula \emph{is} satisfiable, then this first semantic feature is consistent with the scenario and we can move on.
To avoid reasoning about this feature again, we \textit{condition} the ego position expression tree by replacing it with the actual position value from the label. 
This conditioning will apply as we now check the consistency of the next feature in our order, namely the ego's heading.
Translating the second expression tree to SMT yields the formula
\[
e_x{=}l_{e_x} \land e_y{=}l_{e_y}\land(e_h{=}\texttt{roadDir}(e_x, e_y))\land(e_h{=}l_{e_h})
\]
which we call $\varphi_{2} \vert \varphi_{1}$ to indicate the conditioning.
Note that the ego position's expression tree, involving the \texttt{On} predicate, is no longer present, simplifying the formula: this is the essence of our incremental approach.
We then solve the new SMT formula, and repeat this modular feature translation and incremental validation process until either all the features in the label are validated, or a feature is invalidated. 
If all the features are valid, then as a final step we condition all the features as in the label and check whether all the explicit constraints in the \scenic{} program (e.g. line 3 in Fig.~\ref{fig:scenic_program}) are satisfied.
If so, the label matches the scenario; otherwise, it does not.  

%% file: sections/background.tex
\section{Background}\label{sec:background}

\subsection{\scenic: A Scenario Description Language}\label{sec:scenic}
\scenic~\cite{scenic, scenic-journal} is an object-oriented probabilistic programming language designed for \textit{modeling} and \textit{generating} scenarios.
Because \scenic{} is probabilistic, a single \scenic{} program can model \textit{abstract} scenarios like ``bumper-to-bumper traffic'' that cover a wide range of environments, parametrized by semantic features such as weather, time of day, objects' positions, orientations and color.
A \scenic{} program defines a distribution over \emph{scenes}, where a scene is an environment configuration at one point in time (since in this paper we consider only static scenarios) defined by assigning concrete values to the semantic features.
A scene is thus equivalent to a \emph{label} as used in Sec.~\ref{fig:overview}, and we use the terms interchangeably.
The ability of a \scenic{} abstract scenario to generate many scenes will be useful to us when querying, since we can search for data matching a broad description of a scenario.
For the rest of the paper, we refer to abstract scenarios as scenarios for brevity.

To aid in scenario modeling, \scenic{} supports an intuitive syntax to specify complex geometric relations among objects: for example, in Fig.~\ref{fig:scenic_program} we saw the \texttt{on} construct for generating a point uniformly at random within a region, and the \texttt{ahead of} construct for placing one object relative to another.
In addition, \scenic{} enables a user to impose declarative constraints on the distribution of scenes, as in line 3 of Fig.~\ref{fig:scenic_program}, using a \texttt{require} statement.
Here, the constraint succinctly captures the user's intention for the scenario without requiring an explicit description of parts of the road that are sufficiently far from intersections. An execution of \scenic{} program \textit{samples} a scene that must satisfy any \texttt{require} statements specified in a \scenic{} program.

In order to use constructs like \texttt{road} and \texttt{intersection} in Fig.~\ref{fig:scenic_program}, \scenic{} requires basic geometric information about the road network being simulated.
More sophisticated scenarios involving details of the infrastructure (e.g. layout of lanes and crosswalks) require correspondingly detailed maps.
\scenic{} is able to read standard map formats including those provided with the datasets used in this paper.

\subsection{Satisfiability Modulo Theories (SMT)} \label{sec:smt}
The satisfiability problem is the question of whether a propositional formula has a solution, i.e., whether there is an assignment to the Boolean variables that makes the formula evaluate to true.
Over the last three decades, satisfiability solvers have made tremendous progress despite the theoretical hardness of the problem, scaling to formulas with billions of variables~\cite{DBLP:series/faia/SilvaLM09}.

To support problems that involve, e.g., arithmetic, 
SMT solvers ask for a given a first-order formula $\varphi$ over a set of variables and a fixed theory\footnote{A theory fixes the interpretation of the operators, e.g. $+$ meaning addition and $|x|$ denoting the absolute value of $x$.} whether there is an assignment to the variables that makes the formula evaluate to true~\cite{barrett-smtbookch09}. 
% For example, in the theory of real arithmetic, the formula
% $(x^2 + y^2 > 1) \land (x + y < 1)$ is satisfiable: consider, for example, $\langle x \mapsto 0, y \mapsto -2 \rangle$. The formula 
% $(x^2 + y^2 > 1) \land (x+y<1) \land (x > 0) \land (y > 0)$ is not satisfiable.
In this paper, we use a fragment of the theory of quantifier-free nonlinear real arithmetic, with formulas generated by the following grammar:
\begin{align*}
   &  e ::= x \mid a \mid e + e \mid e \times e \mid -e \mid sin(e) \mid arctan(e) \\
   &  c ::= e < 0 \mid e \leq 0 \mid c \land c \mid c \lor c \mid \neg c 
\end{align*}
Here, $x$ is a real-valued variable, $a$ a real-valued constant, and all operators have their standard meanings.
The satisfiability problem for such formulas is undecidable in general (due to the presence of trigonometric functions)~\cite{DBLP:journals/jsyml/Richardson68}.
However, solvers such as dReal~\cite{dreal} can either prove unsatisfiability or return a variable assignment which \emph{approximately} satisfies the formula (in a suitable formal sense; see \citet{dreal} for details).
Thus, in this paper, we do not consider the question of decidability further.

%% file: sections/problem.tex
\section{Problem Statement}\label{sec:problem}
%\subsection{Terminology}

Let a \emph{label} $l$ consist of (1) a set of objects $O$, (2) a set of semantic features $F$, and (3) a function $s \colon O \times F \rightarrow V$ which maps an object's semantic features to concrete values. The domain of values $V$ can include real numbers, integers, categorical values (for examples, see Sec.~\ref{sec:overview}) and the special value $\bot$, which indicates that the object does not have the corresponding semantic feature.
Let $\semantics{P}$ denote the \emph{support} of a \scenic{} program $P$, i.e., the set of labels that can be generated by $P$. 
Then, the problem to solve could be: 

\textbf{Problem P0}: 
\emph{Given a \scenic{} program $P$ and a label $l$, is $l \in \semantics{P}$?}, i.e., is the probability (density) of $l$ under the distribution defined by $P$ greater than 0?

However, the above problem statement is too strict for our purposes. In particular, it may answer "no" for a label $l = (O, F, s)$ which is not in $\semantics{P}$ even if an essentially equivalent label $l' = (O', F', s')$ \emph{is} in $\semantics{P}$.
This can happen in two ways:

(1) The semantic feature spaces of $l$ and $l'$ may differ, i.e., $F \neq F'$. For instance, a car's color and model might be included in $F$ but not in $F'$. In such cases, we will only consider features in $F \cap F'$ as being relevant to our query.

(2) More importantly, $l$ may contain multiple objects, and the correspondence between $O$ and $O'$ is unknown; in fact, we may want to consider $l$ as matching $P$ even if it contains additional objects not having any counterpart in $P$ (for example, we may want a program for ``two perpendicular cars'' to match whenever two such cars exist, even if there is a third car in the vicinity).

% (3) Lastly, some of the labels which are "close" to matching the program will get rejected because they are technically not in $\semantics{P}$. In other words, the query may not be \textit{robust} to some noise in the label. \sebastian{See email}

Therefore, we generalize \textbf{P0} as follows. 

\textbf{Problem P1}: 
Let a label $l$ \emph{match} a program $P$ if there exists $l' = (O', F', s') \in \semantics{P}$ and an injective\footnote{If we want to require an \emph{exact} match in the sense that $l$ does not contain any objects beyond those defined in $P$, we can also require $C$ to be surjective. Our algorithm extends to this case with only trivial changes.} function $C \colon O' \rightarrow O$ such that $\forall o' \in O'$,  $s'(o', F\cap F') = s(C(o'), F\cap F')$. Here the function $C$, which we call the \emph{object correspondence}, maps each object in $P$ to a distinct object in $l$. Then:
{\it Given program $P$ and label $l$, determine whether $l$ matches $P$}. 

To account for any possible noise in the labels, one may modify the \scenic{} program to account for these noise with respect to particular semantic features of interest.

\begin{figure}[t]
  \centering
  \includegraphics[width=0.40\textwidth]{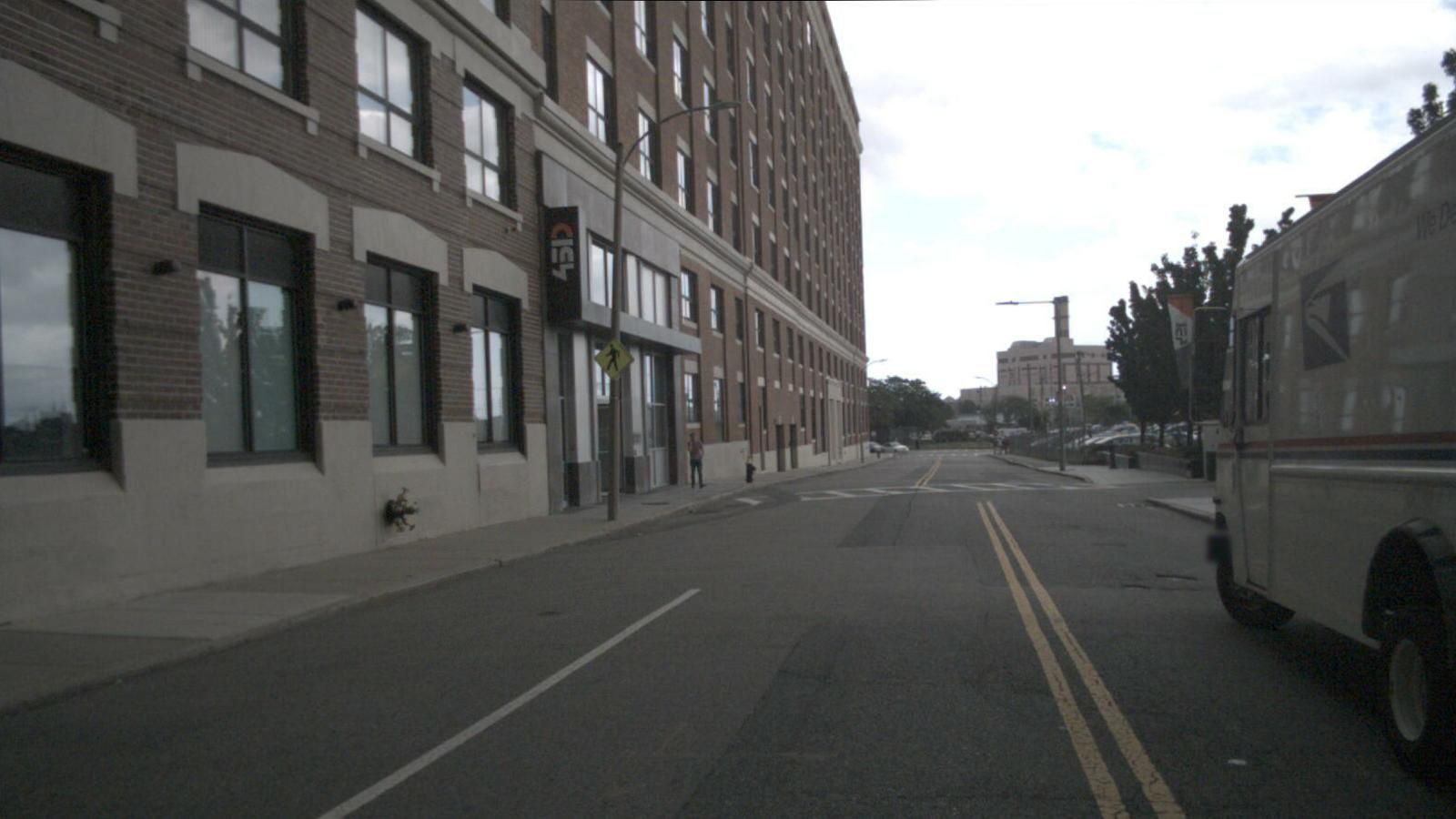}
  \hfill
  \includegraphics[width=0.40\textwidth]{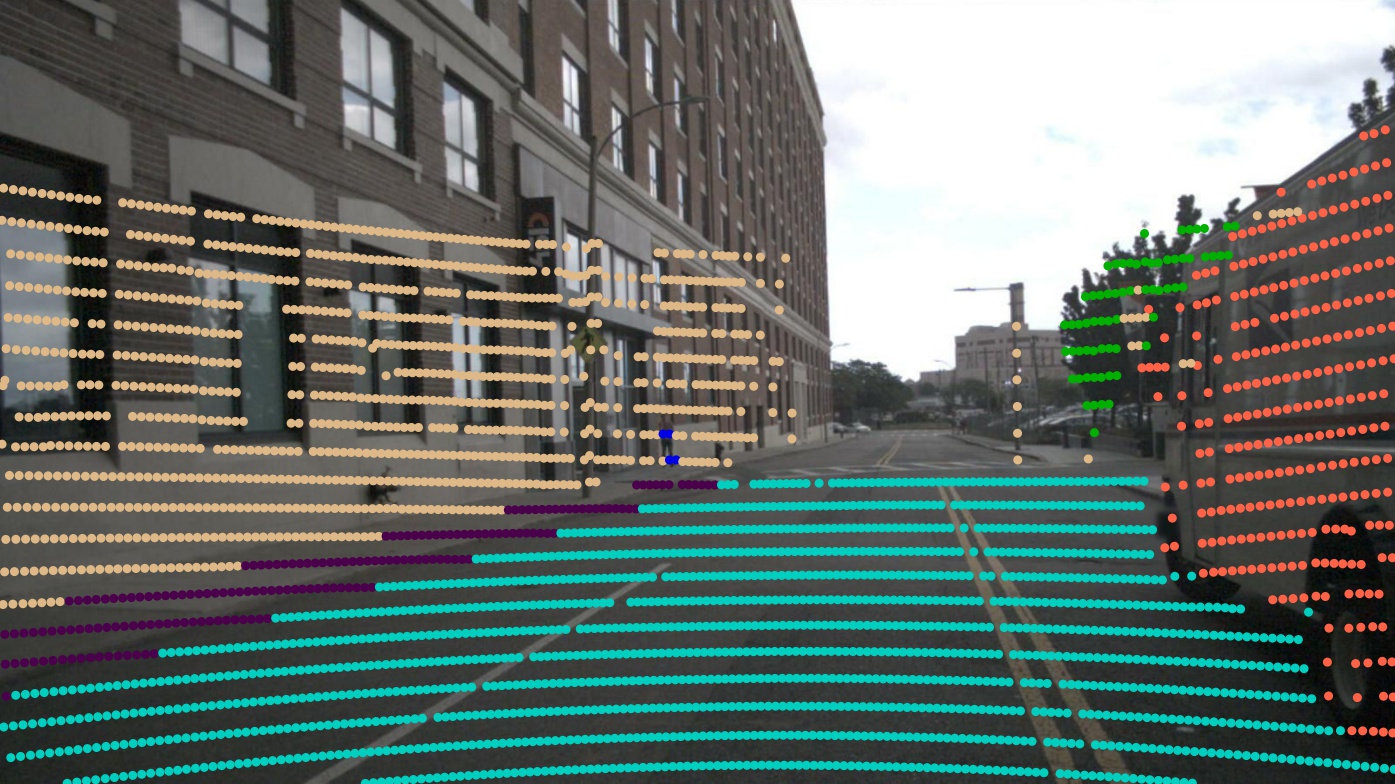}
  \caption{The benefit of querying a label is that we can retrieve corresponding sensor data of multiple types. Here, our algorithm finds both an RGB image and a segmented 3D LiDAR point cloud corresponding to a single label.}
  \label{fig:rgb_lidar}
\end{figure}

Our problem formulation can be applied to validate the behaviors of models for different perception tasks using various sensor types (e.g. LiDAR segmentation,  RGB detection) across synthetic and real data. Because we query the labels, our problem formulation broadly applies to querying various labelled sensor types such as 2D RGB, radar, 3D LiDAR point clouds, etc. For example, it is possible that for the same label, data from multiple sensors exists. Fig.~\ref{fig:rgb_lidar} shows 2D RGB image and 3D LiDAR point cloud data that corresponds to the same label. As we query labels rather than the raw sensor data, we can query both sensor types. 

\paragraph{\textbf{Assumptions on the Labels and their Sources}}\label{sec:label_assumption}
To use our methodology in the context of validating sensor data, the sensor data must be adequately labelled. Specifically, at minimum the labels must contain information about the position, orientation, and type (e.g. pedestrian, car, bicycle) of each object. Furthermore, the label information needs to be aligned with the map information (i.e. share the same coordinate system and global orientation).
Labels may also include more diverse global or object-specific semantic features, allowing more expressive descriptions of scenes. For example, global semantic features could include weather and time of day (which determines the location of the sun and, therefore, of shadows). Object semantic features could include color and physical dimensions.

To query AV scenarios involving road geometry, we require corresponding map information: specifically, polygons for regions such as lanes, roads, and intersections, as well as the traffic flow directions within those regions.
This information is needed to interpret \scenic{} constructs such as \texttt{on road}, seen in our example above.
More detailed information may be needed depending on the scenes one may model. For example, if one wishes to model scenes involving cyclists on a bicycle lane in a \scenic{} program, a polygonal region and traffic flow direction for bicycle lanes in the map must be provided. For more details, see Sec.~\ref{sec:scenic}. 

The above two assumptions are needed for querying labels. However, for our motivating application of validating a failure scenario of a perception model, the real labelled dataset should also contain relevant ground truth labels to evaluate the perception task (e.g. segmentation, detection). For example, if the task is 3D detection, the 3D bounding boxes should be included in the label. Then, once we retrieve a matching subset of real data, the perception model can be evaluated on this subset with the relevant ground truth labels. 

There are open-source datasets that satisfy the two assumptions for querying and also provide ground truth labels sufficient for various perception tasks, such as nuScenes~\cite{nuscenes} and Argoverse~\cite{argoverse}; we used nuScenes for our experiments.

%% file: sections/methodology.tex
\section{Methodology}\label{sec:method}

Given a label and a \scenic{} program, the key idea behind our approach is to translate the program to an SMT formula that is satisfied if and only if the label matches the program.

\subsection{Monolithic Approach}\label{sec:monolithic}
The basic set-up of the SMT formula, as outlined in Sec.~\ref{sec:overview}, defines variables for all objects and semantic features in the intersection of program and the label.
There are three aspects to the SMT formula; the first aspect of the formula maps objects in the label with objects in the program (which we referred to as \textit{correspondence} in Sec.~\ref{sec:problem}), the second aspect describes the constraints over all the semantic features in the program as dictated by the semantics of \scenic{}, and the third aspect asserts that the semantic feature values observed in the label satisfies those corresponding constraints in the program.

The \scenic{} compiler represents a compiled \scenic{} program as an \emph{expression forest}.
Each semantic feature is the root node of an \emph{expression tree} appearing in the forest, which captures the semantics of how the feature's value is derived from the values of other features and random parameters.
To encode the semantics of the \scenic{} program, we walk the expression forest, generating SMT equivalents of each of the nodes.
For example, the \texttt{On} node in Fig.~\ref{fig:expression_tree} is encoded by a set of constraints enforcing that the variable representing the ego's position must lie in the \texttt{road} region\footnote{This is done by triangulating the region, and using a disjunction of linear inequalities to assert that the position lies in one of the triangles. See Appendix~\ref{appendix:encoding} for more details.}.
All \scenic{} constructs\footnote{Our implementation handles a large fragment of \scenic{}, supporting its built-in operators and functions. However, it does not support the inclusion of arbitrary Python code, which \scenic{} allows in some contexts.} can be encoded as real arithmetic constraints fairly easily, following the \scenic{} semantics outlined in \cite{scenic}. The details of our SMT encoding and the fragment of \scenic{} we support are explained in Appendices~\ref{appendix:encoding} and ~\ref{appendix:fragment}, respectively.

This method yields a sizeable SMT formula, which is difficult to solve beyond toy examples. Therefore, we consider a modular approach outlined below.

\subsection{Modular and Incremental Approach} \label{sec:modular_approach}
Our approach is formalized as Algorithm~\ref{algorithm}. The input to the algorithm is a \scenic{} program $P$ and a label $l$, and the output is ``Yes'' if $l$ matches $P$ and otherwise ``No.''
The algorithm has three main steps: (1) dependency analysis of objects and their features, (2)~incremental translation and validation of the program to a series of SMT formulas, and (3)~validation of hard constraints.
We now discuss each step in detail.

\newcommand{\badcorrs}{\textit{badOCs}}
\newcommand{\nextfeatures}{\textit{nextFeatures}}
\begin{algorithm}[t]
\caption{Our proposed query algorithm}
\label{algorithm}
\textbf{Input} : A \scenic{} program $P$, a label $l$, and visible distance \\
\textbf{Output}: Does $l$ match $P$? 
\begin{algorithmic}[1]
\STATE $\textit{EF} \gets \textit{Compile}(P)$ // get expression forest
\STATE $\text{SortedFeatures} \gets \textit{AnalyzeDependencies}(\textit{EF}, l)$
\STATE $\badcorrs \gets \emptyset$ // partial correspondences that don't work
\FOR {all possible object correspondences $C$}
\IF {$C$ extends a correspondence in \badcorrs}
\STATE \textbf{continue}; // skip this correspondence
\ENDIF
\STATE $\textit{failed} \gets \textit{false}$
\FOR {$\nextfeatures\in\text{SortedFeatures}$}
\STATE $\varphi \gets$ \textit{TranslateSMT}(\nextfeatures, $l$, $C$, \textit{EF})
\IF{$\textit{Satisfiable}$($\varphi$)}
\STATE \textit{Condition}(\nextfeatures, $l$, $C$, \textit{EF})
\ELSE
\STATE \textit{failed} $\gets$ \textit{true} \\
\STATE $\textit{Uncondition}(\textit{EF})$ // reset forest
\STATE add the used part of $C$ to \badcorrs
\STATE \textbf{break}
\ENDIF
\ENDFOR
\IF{($\textbf{not}$ \textit{failed}) $\textbf{and}$ \textit{SatisfiesHardConstraint}{($l$)}}
\STATE \textbf{return} \textit{Yes}
\ENDIF
\ENDFOR
\STATE \textbf{return} \textit{No}
\end{algorithmic}

\end{algorithm}

\subsubsection{Object/Feature Dependency Analysis}\label{sec:dependency_analysis}
\begin{figure}[t]
    \centering
    \includegraphics[width=\linewidth]{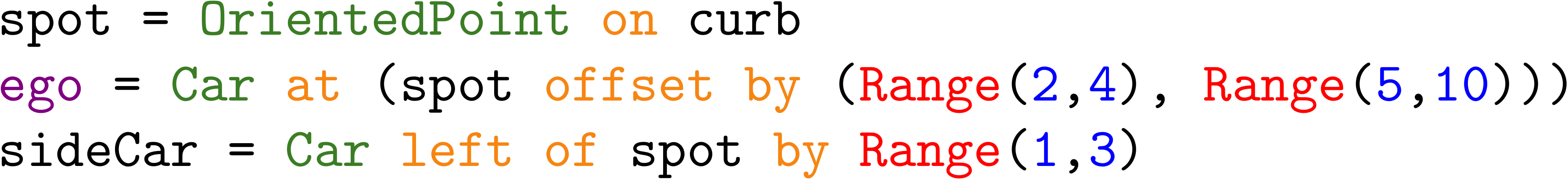}
    \caption{A \scenic{} program with an intermediate variable, \texttt{spot}, shared between position features of \texttt{ego} and \texttt{sideCar} objects.}
    \label{fig:shared_variable}
\end{figure}

% \sebastian{Dependencies seem to be parameterized by the label}
A key feature of our approach is to exploit dependency structure in the \scenic program, specifically its compiled expression forest, to split the monolithic SMT query into smaller parts. 
We define two types of dependencies in the expression forest: \textit{dependent} and \textit{jointly dependent}. If an expression tree of a feature, $X$, has a reference to another feature, $Y$, then $X$ is \textit{dependent} on $Y$.
Such dependencies are \textit{acyclic} because a feature that is specified first in \scenic{} cannot reference an object defined afterwards. If two or more feature expression trees share internal nodes which are not features, then those features are \textit{jointly dependent}. These shared node(s) are intermediate (i.e., unobserved) variables which are not part of the scene. Therefore, to check whether there exists a feasible value for the intermediate variables, jointly dependent features must be considered in the same SMT query. For example, Fig.~\ref{fig:shared_variable} shows a \scenic{} program describing a distribution of scenes with two cars positioned in parallel, adjacent to a spot uniformly randomly selected from a curb region. In this case, the ego and side car's position features are \textit{jointly} dependent on the intermediate variable, \texttt{spot}. Note that this \texttt{spot} is an intermediate variable introduced in the \scenic{} program. Hence, we need to encode both cars into the SMT query to check if there exists a value of \texttt{spot} which satisfies the constraints with both of their positions. 
Note that for the following dependency analysis among semantic features, as we stated in Sec.~\ref{sec:problem}, we are only analyzing semantic features that exist both in the \scenic{} program and the label. Hence, this dependency analysis takes as inputs, both the expression forest of the given \scenic{} program and the label (line 1 of Alg.~\ref{algorithm}). 

Once we identify these dependency relations across all object features, we sort the features in dependency order, giving jointly dependent features the same rank in the order (line 2 in Alg.~\ref{algorithm}).
For example, analyzing the expression forest in Fig.~\ref{fig:expression_tree}, with no jointly dependent features, yields the order [\{\texttt{ego} position\}, \{\texttt{ego} heading\}, \{\texttt{otherCar} position\}, \{\texttt{otherCar} heading\}]. For the \scenic program in Fig.~\ref{fig:shared_variable} with jointly dependent features, our analysis outputs [\{\texttt{ego} position, \texttt{sideCar} position\}, \{\texttt{ego} heading\}, \{\texttt{sideCar} heading\}]. 

\subsubsection{Modular and Incremental SMT Translation}\label{sec:modular}
Given the sorted feature dependency list obtained as above, we translate only one feature expression tree (or multiple trees, if jointly dependent) at a time, in the order of dependency (line 8-9 in Alg.~\ref{algorithm}).
Checking the resulting SMT query, if the formula is unsatisfiable then it is impossible for the current features to take on their observed values and so the label does not match the program (line 12-13).
If instead the formula is satisfiable, then the observed values are feasible given the semantics of the program, and we need not consider the current features further: we \emph{condition} (i.e. equate) the expression forest on their observed values, substituting the values in for their expression trees (line 10-11).
Then we move on to the next feature(s) in the dependency order and repeat.
If a previously-checked feature is referenced by a later expression tree, we do not need to encode it again, since it now has a constant value.
This modular approach can significantly simplify the generated SMT queries: for example in Fig.~\ref{fig:expression_tree}, instead of one query encoding the entire forest, we have one query for each of the 4 top nodes encoding only the nodes directly below it in the order of dependency (refer to Sec.~\ref{sec:overview} for more detail).

We remark that our modular translation requires fixing the correspondence of label and program objects \emph{a priori}.
As seen in the outermost loop of Alg.~\ref{algorithm} (line 4), we currently brute-force enumerate all possible combinations, with one refinement: if a partial correspondence is already enough to make the SMT query unsatisfiable, we can exclude all further correspondences extending it.
In particular, when a feature of object $O$ fails the SMT check under correspondence $C$, the part of $C$ consisting of all objects up to and including $O$ (in dependency order) will also yield unsatisfiability.
So we maintain a set of partial correspondences known not to work (line 15) and skip any correspondence which extends one of them (line 6).
Our experiments show that this approach scales to a reasonable number of objects.

Also, regarding encoding polygonal regions of the map (e.g. lanes, roads, intersections), they can be quite sizeable to encode. For example, nuScenes, which is the dataset we used for our Experiment, provides a map information of Boston city. Encoding the entire road network of Boston may be impractical since we are interested in the scene around the ego vehicle, which determines the reference viewpoint of sensors (e.g. camera, LiDAR). Hence, we set ego's visible distance as a hyperparameter, and only encode the polygonal regions of the map within that radius of visible distance (line 1). 

% , and potentially more if objects have distinguishing features (such as their color).

\subsubsection{Hard Constraint Validation}
Finally, if the \scenic{} program contains any \texttt{require}  statements encoding hard constraints (see Sec.~\ref{sec:scenic}), we need to check that these are satisfied by the label.
After all features have been validated and conditioned on their observed values, we simply check that the \texttt{require} constraints all evaluate to true. 
Note that if any \texttt{require} constraints were jointly dependent (see Sec.~\ref{sec:modular_approach}) with any feature, for soundness we would have to encode the constraints into the SMT query for that feature.
Since the \scenic{} compiler currently does not generate expression trees for requirements, we instead assume that the program does not have any such joint dependencies (as mentioned in our \scenic{} fragment~\ref{appendix:fragment}).
This assumption holds for the vast majority of the scenarios in the \scenic{} distribution~\cite{scenic}. 

Our proposed query algorithm is sound as stated in Theorem 1 whose proof is provided in Appendix~\ref{appendix:proof}.

\begin{theorem}
Given a label, a \scenic{} program, the \scenic{} query algorithm outputs \texttt{True} if and only if the label \textit{matches} the program as defined in Sec.~\ref{sec:problem}; otherwise, it outputs \texttt{False} (assuming the underlying SMT solver correctly answers all queries).
\end{theorem}

%% file: sections/experiment.tex
\section{Experiments}\label{sec:experiment}
Recall, in Sec.~\ref{sec:intro}, we proposed the query problem to address the scenario validation issue. Once our algorithm queries a labelled, real dataset with a scenario encoded in \scenic{} and retrieves a matching subset, then validating a perception model's behavior on that subset is straightforward (assuming the dataset also contains the relevant ground truth labels for the perception task: see Sec.~\ref{sec:label_assumption}). In fact, the most time-consuming aspect in this validation process is the query.
Thus, to evaluate How useful our algorithm is, we ask the following questions:
\begin{figure*}[t]
    \centering
    \includegraphics[width=0.245\linewidth]{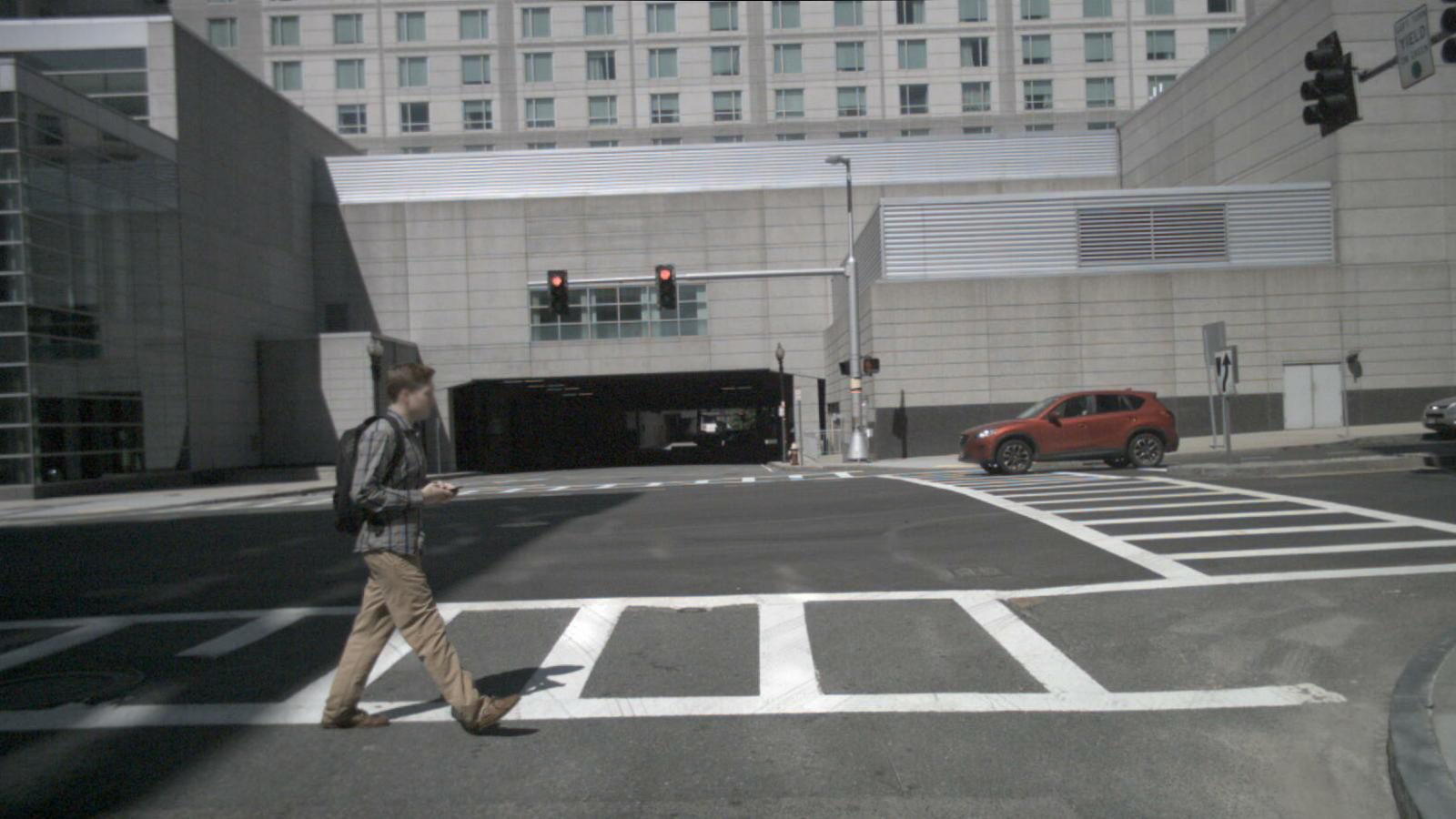}
    \hfill
    \includegraphics[width=0.245\linewidth]{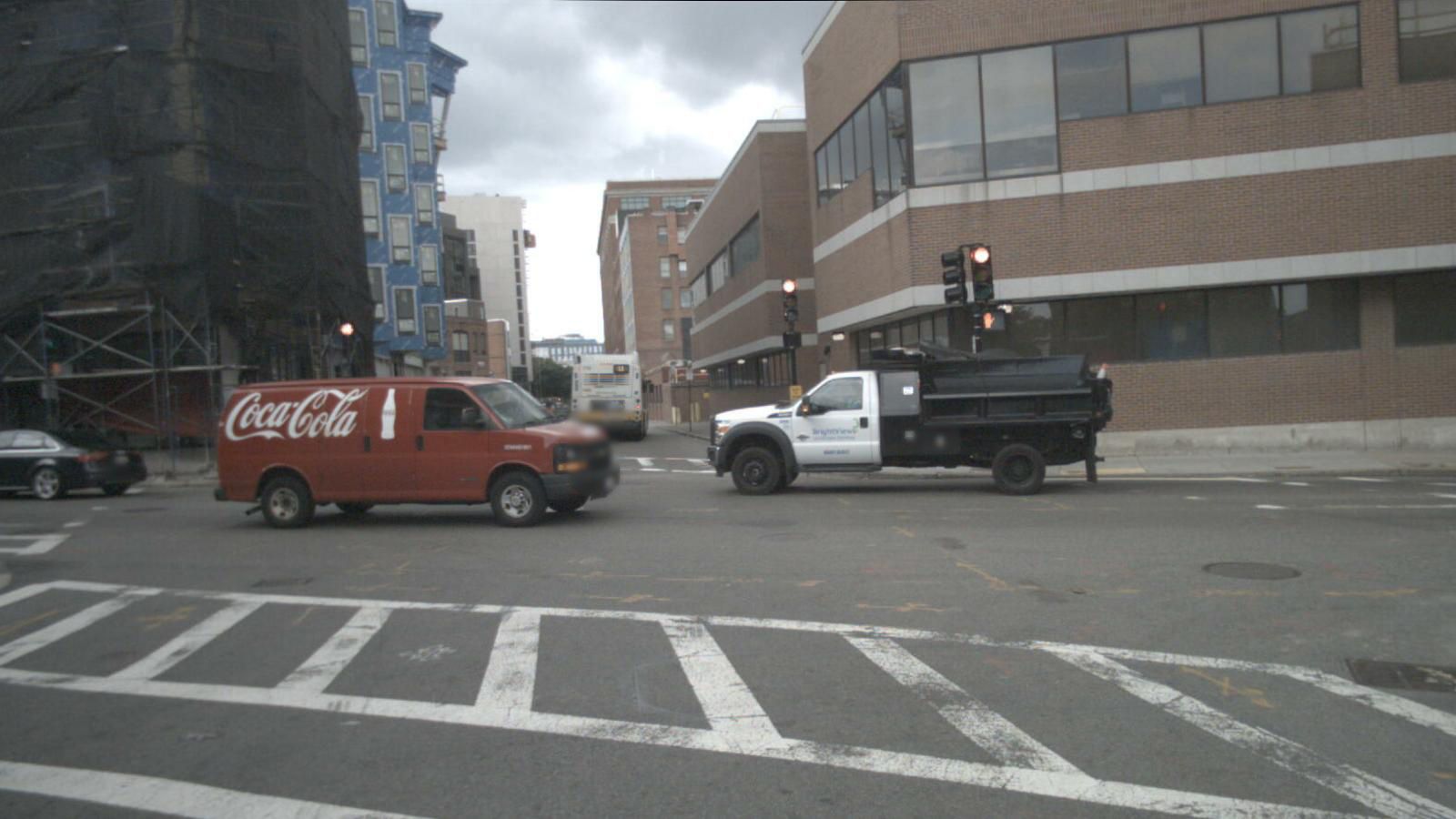}
    \hfill
    \includegraphics[width=0.245\linewidth]{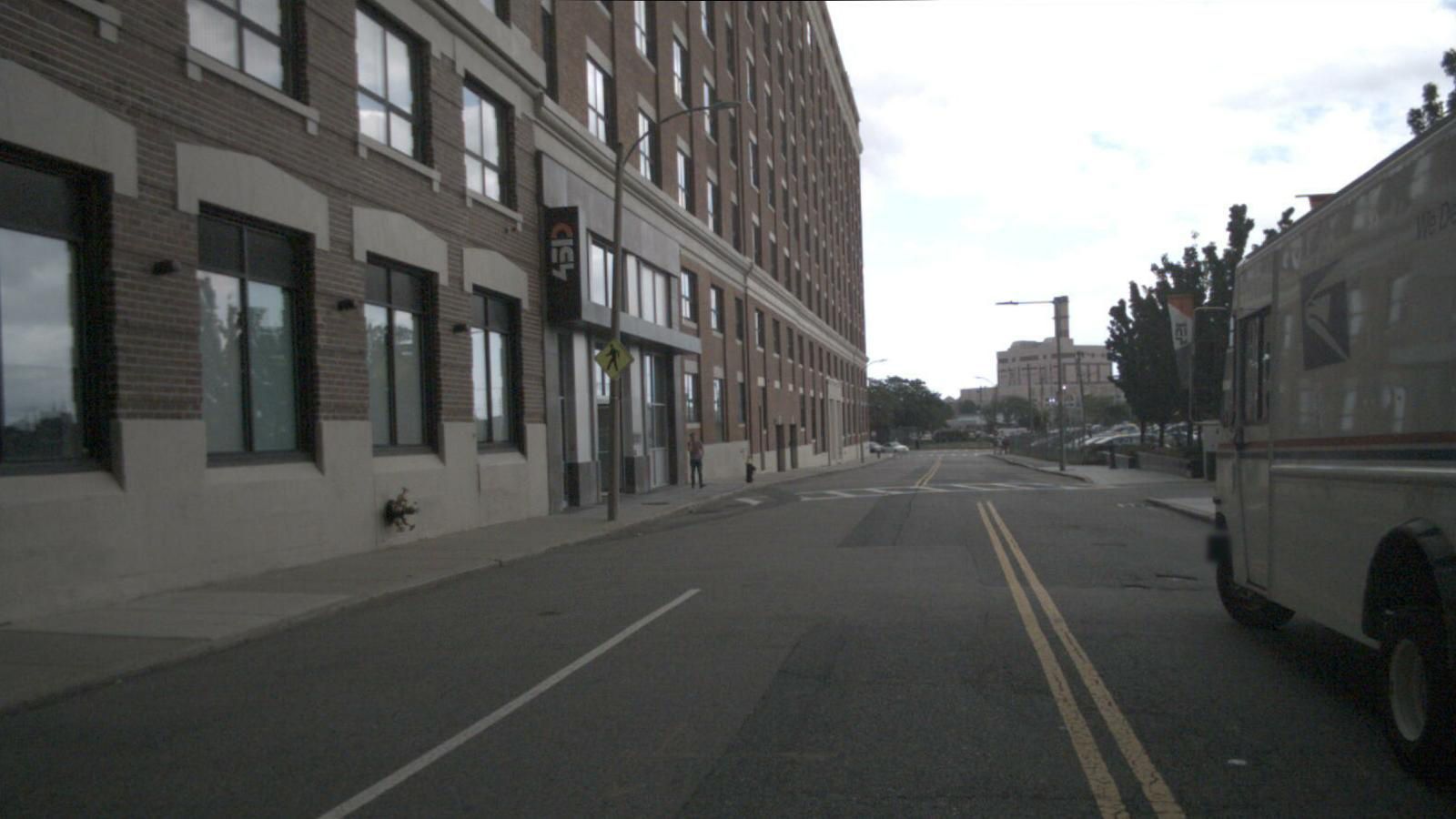}
    \hfill
    \includegraphics[width=0.245\linewidth]{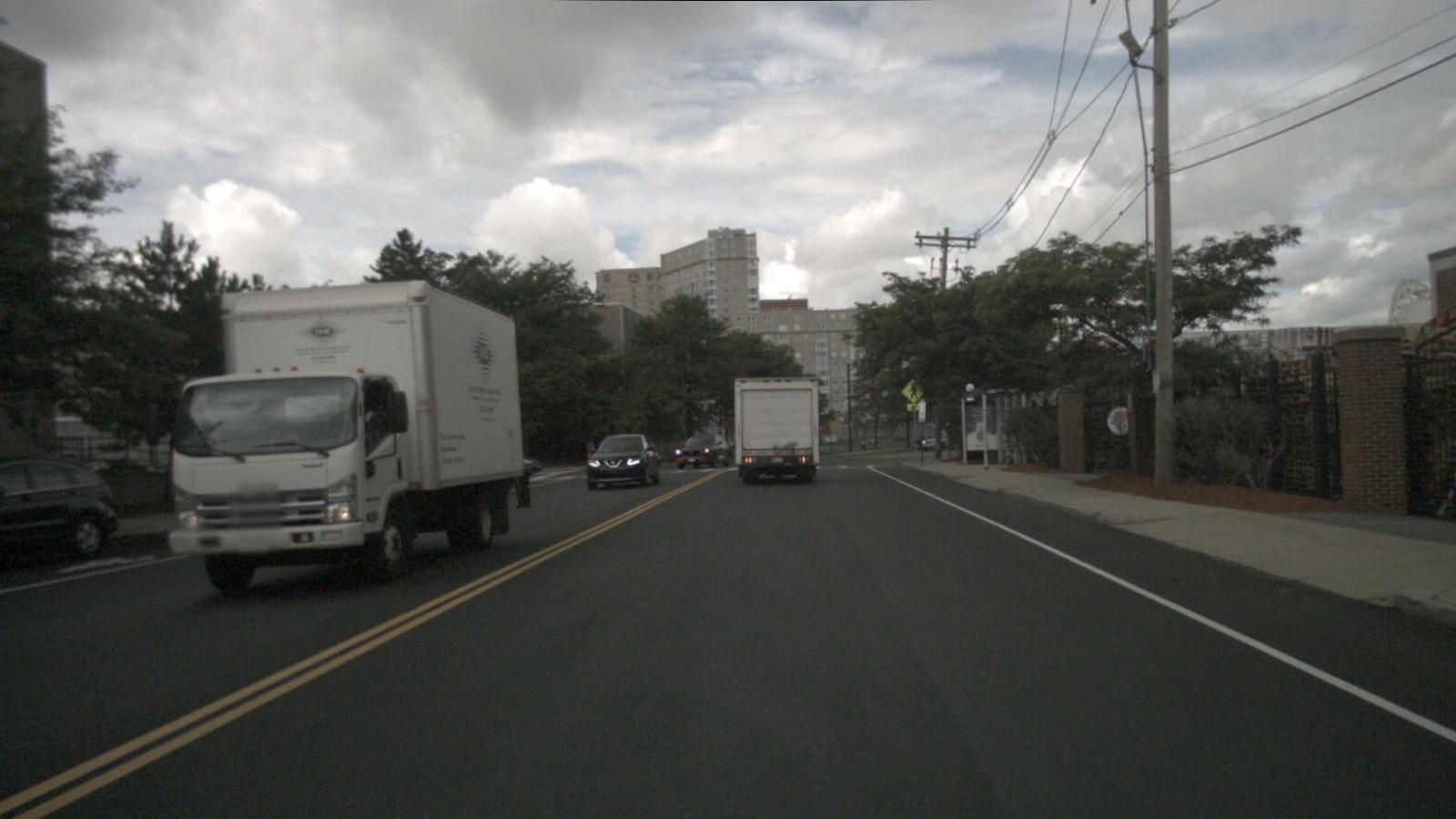}
    \caption{Matching images for Scenarios 1 through 4 (left to right), queried using our algorithm.}
    \label{fig:queried_images}
\end{figure*}

\begin{enumerate}
\item Given a \scenic{} program and a real labelled dataset, does the algorithm efficiently find the matching data points?
\item Does the output of the algorithm correspond with the intuitive notion of scenario matching?
\item How does the algorithm scale with scenario complexity, in terms of number of agents and program structure?
\end{enumerate}
The second question is important to address since it directly relates to the \textit{interpretability} of the scenario validation process which is crucial for debugging. Therefore, we need to check whether this algorithm operates in a manner intuitive to humans.

To answer these questions, we conducted two different experiments.
First, our efficacy experiment answers the first two questions. In a nutshell, it demonstrates that the formal querying problem we define and solve corresponds well to our intuition of what it means to match an image against a high-level scenario and is efficient in comparison to manual querying. Our second experiment demonstrates that our approach remains feasible even on fairly large scenarios. In both experiments, we used the dReal SMT solver~\cite{dreal} with its default parameter settings and using quantifier-free real arithmetic. For the first experiment, we set the ego visible distance to 50 meters; for the second, we set it to 200 meters to accommodate the larger number of agents.

\begin{table}
{
\centering
\begin{tabular}{*6c}
\toprule
Scenario \#    & 1  & 2  & 3  & 4 & 5  \\ 
\midrule
Matching images (humans) & 42 & 5 & 0 & 2 & 0 \\ 
Matching images (our algorithm) & 58 & 7 & 2 & 2 & 0 \\
\bottomrule
\end{tabular}
\caption{For several scenarios, the number of images identified by 3 human subjects (unanimously) and our algorithm.}
\label{table:human_experiment}
}
\end{table}

% \begin{table}
% {
% \centering
% \begin{tabular}{*3c}    
% \toprule
% Scenario $\#$ & CARLA & nuScenes  \\ 
% \midrule
% Scenario $\#2$ & TBD & TBD
% \bottomrule
% \end{tabular}
% \\
% \caption{For validation of a pre-trained 3D object detection model's performance (measured in mean average precision (mAP)) on Scenario $\#2$ across the synthetic CARLA and real nuScenes 2D RGB images are shown.}
% \label{table:validation_experiment}
% }
% \end{table}

% \begin{table}
% {
% \centering
% \begin{tabular}{*3c}    \toprule
% \# of Agents & Obj Corr. Known & Obj Corr. Unknown \\
% \midrule
% 1 & $0.21 \pm 0.04$ & $0.18 \pm 0.04$ \\ 
% 2 & $0.27 \pm 0.03$ & $0.23 \pm 0.05$\\
% 3 & $0.38 \pm 0.03$ & $0.47 \pm 0.12$\\
% 4 & $0.62 \pm 0.07$ & $1.16 \pm 0.53$\\
% 5 & $0.72 \pm 0.07$ & $1.70 \pm 0.84$\\
% 6 & $0.76 \pm 0.08$ & $2.13 \pm 0.88$\\
% 7 & $0.98 \pm 0.06$ & $7.10 \pm 4.03$\\
% 8 & $1.08 \pm 0.10$ & $4.99 \pm 2.15$\\
% 9 & $1.29 \pm 0.10$ & $8.84 \pm 3.22$\\
% 10 & $1.37 \pm 0.13$ & $9.68 \pm 4.48$
% \bottomrule
% \end{tabular}
% \caption{Average query runtimes (s) of our algorithm for scenarios with various numbers of agents, with and without the object correspondence (see Sec.~\ref{sec:problem}) known \emph{a priori}. Averages and standard deviations are over 10 queries.}
% \label{table:scenic_experiment}
% }
% \end{table}

\subsection{Efficacy Experiment}\label{sec:efficacy_experiment}
% In motivation for developing query algorithm, as visualized in Fig.~\ref{fig:vision}, is to validate a pre-trained AV perception model on given scenario across simulated and real data. The bottleneck in this validation process is to retrieve real data that \textit{matches} the scenario in a given labelled dataset. The proposed query algorithm aims to automate this retrieval process. The purpose of this experiment is to investigate whether (i) this algorithm can efficiently do so in comparison to manual query and (ii) the output of the algorithm matches intuitive notion of scenario matching.

\paragraph{Setup}
There is no baseline or benchmark to which we can compare our algorithm since there are no existing algorithms for the problem of querying with a formal scenario description, or open-source autonomous driving image datasets that provide detailed formal scenario descriptions with which to test our algorithm. Hence, we asked 3 human participants to manually query a set of images with five different scenarios and then compared their results with the outputs of our algorithm. 
We asked each participant to select 5 subsets of images matching more detailed versions of the natural language descriptions of our test scenarios below. 
To acquire the most accurate queried subsets, we kept only the images which all 3 humans agreed matched for each scenario. We then compared these subsets with those returned by our algorithm.

\paragraph{Scenarios}
We used five scenarios, involving 2--4 agents and a variety of realistic traffic situations.
Here we provide natural language descriptions.
Several example matching images are shown in Fig.~\ref{fig:queried_images}, and \scenic{} encoding of these scenarios are shown in Fig.~\ref{fig:efficacy1}--\ref{fig:efficacy5}.
\begin{enumerate}
\item
A pedestrian in an intersection facing nearly perpendicularly or towards the ego.
\item
Two vehicles in an intersection, travelling perpendicular to the ego.
\item 
A rare, hazardous situation, where the ego vehicle is driving against traffic and another vehicle is visible within 10 meters.
\item
Four vehicles in a typical situation on a two-lane road, with two vehicles going in each direction.
\item 
A cut-in scenario where a car in the adjacent lane to the right cuts in front of the ego.
\end{enumerate}

\paragraph{Data}
We use a selection of RGB images from nuScenes~\cite{nuscenes}. nuScenes provides the map of Boston where the images were collected; our scenarios used map information about intersection and lane regions as well as traffic flow directions.
nuScenes labels contain three semantic features per object: its position, heading, and class. The object classes include vehicles, pedestrians, and static objects such as traffic cones.
nuScenes also defines a special object class, ``ego vehicle'', indicating the reference viewpoint, meaning the camera for image collection is mounted on it.
As we will describe below, our experiment required humans to identify matches between programs and labelled images; to avoid objects being missed by the humans due to visual occlusion, we filtered out images containing more than 4 objects. After filtering, we randomly selected 700 images from the subset, which we believe is a reasonably large dataset for human participants to manually query. 

Note that the semantic features in the label and the program determines the matching real data. In our experiment, we limited our query to use only semantic features included in the nuScenes labels, namely position, heading, and type for each object.
%However, for the scenario validation with a perception model, depending on the data sensor type (e.g. RGB, LiDAR, radar) it uses, other semantic features may also affect the model's performance such as weather, time of day (which affects the location of the sun), etc. As to which level of abstraction over the semantic features is expressive enough for this validation process needs to be further investigated. 

% Second, we demonstrate the validation of a scenario across the simulated and the real domains. We selected Scenario $\#2$ since it had the most queried images without errors in the labels which is a limitation we cover in the results. Then, we generated scenes with the \scenic{} program and rendered them with CARLA simulator~\cite{carla}, the same number of matching synthetic 2D RGB images as the number of queried images using Scenario $\#2$. Finally, we compared the performance of a pre-trained perception model [CITE], which is trained on real 2D RGB nuScenes data for 3D object detection task, to validate whether the perception model behaves differently in the same scenario but across two different domains. 

% \begin{figure}
%   \centering
%   \includegraphics[width=0.49\linewidth]{figures/against_traffic1.jpg}
%   \hfill
%   \includegraphics[width=0.49\linewidth]{figures/lidar1.jpg}
%   \caption{This is an example scene not identified by manual query for Scenario 3. Two types of sensor data are retrieved by our query algorithm for the same scene, RGB image (left) and segmented 3D LiDAR point clouds (right).}
%   \label{fig:missed_scenes}
% \end{figure}

\paragraph{Results}
First, our results, summarized in Table~\ref{table:human_experiment}, show that our algorithm corresponds with the intuitive notion of scenario matching. For all 5 scenarios, our algorithm correctly returned all images identified by the human participants, and, in some cases, found additional matching images that they missed. For Scenarios 2 and 3, our algorithm found 4 additional images that our participants missed by mistake.
An example missed image for Scenario 3 is shown in Fig.~\ref{fig:queried_images}.
For Scenario 1, our algorithm identified 16 more images than the participants; however, upon visual inspection, 8 of these did not match the scenario description and the remaining were missed by our participants.
Investigating further, we found that these errors were caused by inaccurate labels in nuScenes, e.g.~a pedestrian's position being in an intersection according to the label but being on a sidewalk near an intersection in the image. Our algorithm correctly identified such labels as matching Scenario 1, even though the sensor data disagreed. This illustrates a limitation of our approach in that it hinges on the accuracy of the provided labels.
Finally, for Scenario $\#5$, both the human participants and the algorithm have consensus that there is no matching scenes in the given real dataset. This shows the strength of our algorithm which can also identify "rare" scenarios in a given dataset. Especially, if an AV perception model fails in such rare scenarios in simulation, this finding can guide the real-world data collection process to \textit{systematically} gather specifically such scenarios of data. 
% Overall, the experiment shows that our algorithm is accurate on a variety of realistic scenarios we can encounter on roads. 

The algorithm's runtime over all 700 labels ranged from 7 minutes (scenario $\#1$) to 40 minutes (scenario $\#4$) depending on the complexity of scenarios. The human participant took on average about 1 hour to complete their manual queries on the five provided scenarios. This result demonstrates that our algorithm can replace the arduous task of manually querying matching real sensor data for scenario validation with higher accuracy, provided that the labels are accurate.

\begin{figure}
    \centering
    \includegraphics[width=\columnwidth]{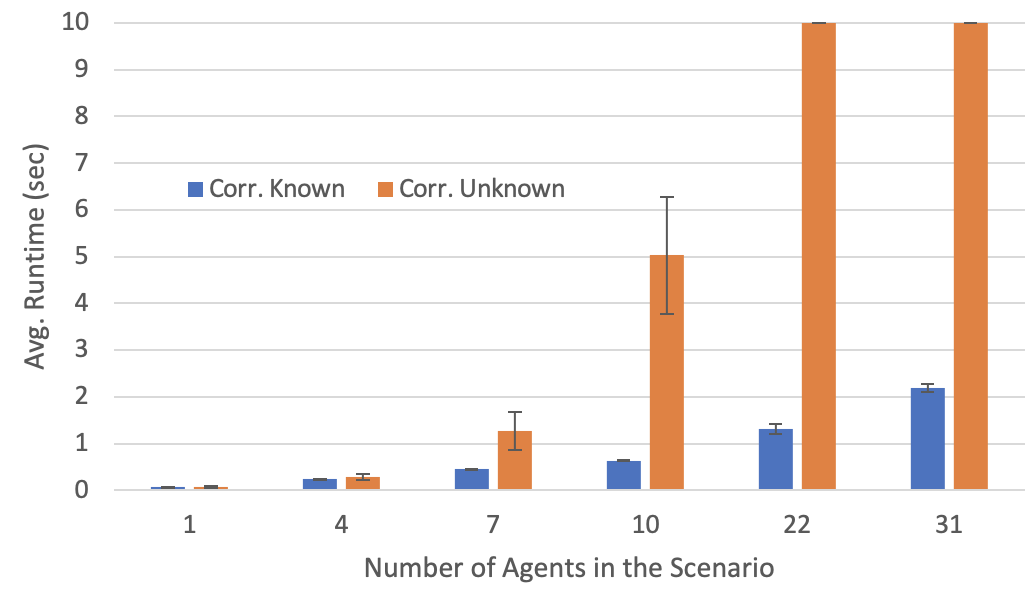}
    \caption{Runtime results for scaling number of agents in the \scenic{} program shown in Fig.~\ref{fig:scalability1}. Two runtimes represent cases when the object correspondence between the program and the label is known versus unknown.}
    \label{fig:scalability1_chart}
\end{figure}

\begin{figure}
    \centering
    \includegraphics[width=\columnwidth]{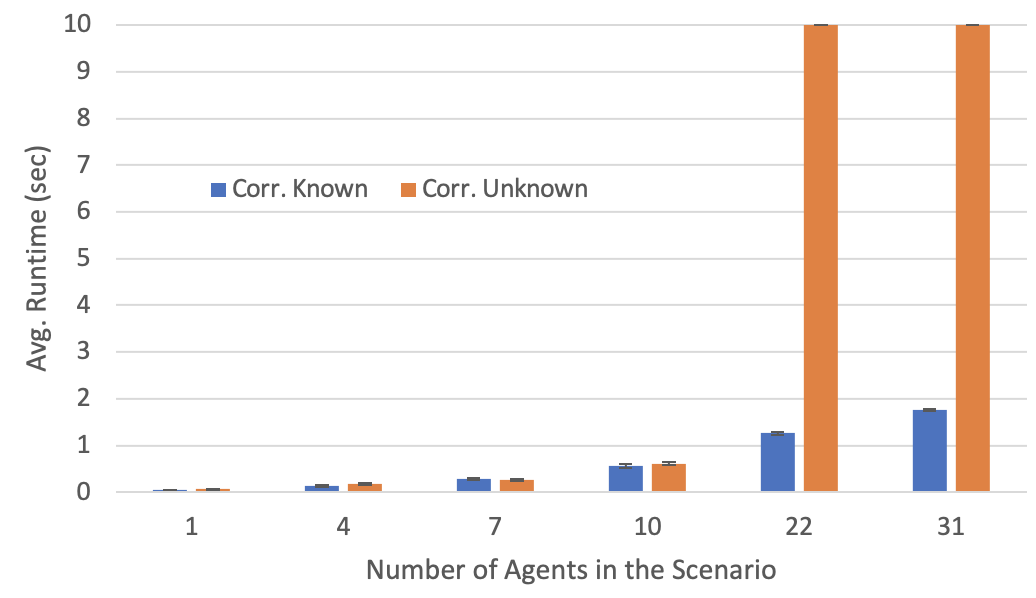}
    \caption{Runtime results for scaling number of agents in the \scenic{} program shown in Fig.~\ref{fig:scalability3}. Two runtimes represent cases when the object correspondence between the program and the label is known versus unknown.}
    \label{fig:scalability3_chart}
\end{figure}

\subsection{Scalability Experiment}\label{sec:scenic_experiment}
\begin{figure}
    \centering
    \includegraphics[width=\columnwidth]{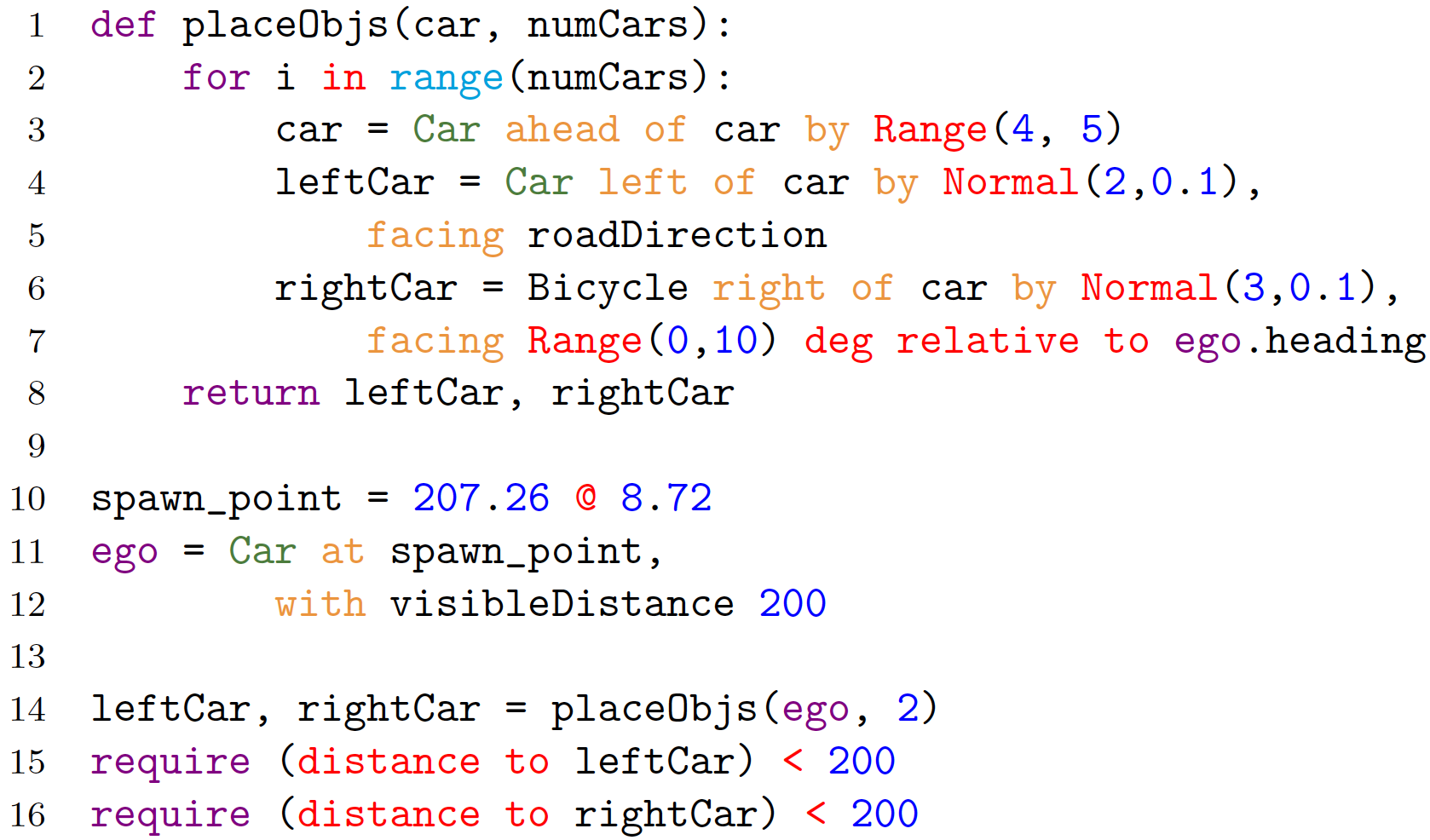}
    \caption{A \scenic{} program used for scalability experiment, modeling bumper-to-bumper traffic. The number of vehicles in the scenario is scaled by increasing the \texttt{numCars}}
    \label{fig:scalability1}
\end{figure}

\begin{figure}
    \centering
    \includegraphics[width=\columnwidth]{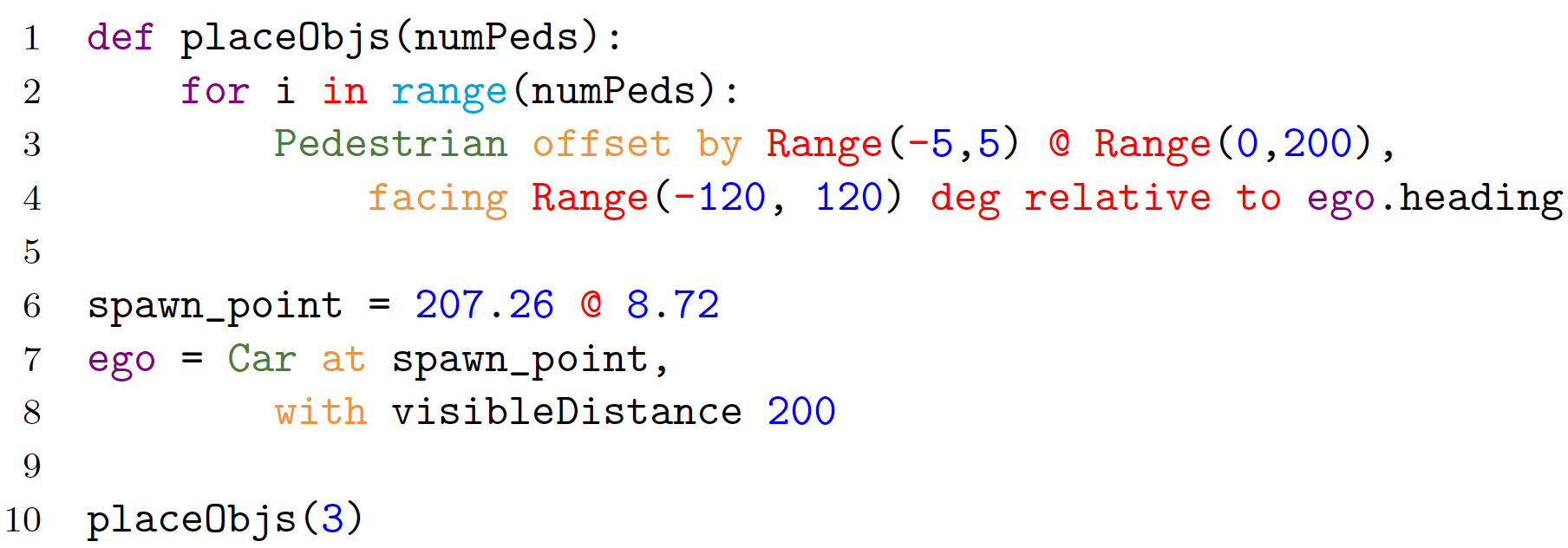}
    \caption{A \scenic{} program used for scalability experiment, modeling a parade scenario where pedestrians are walking on a street. The number of pedestrians in the scenario is scaled by increasing the \texttt{numPeds}}
    \label{fig:scalability3}
\end{figure}

%As aforementioned, in our human experiment, we limited the number of objects in the label to four at max to avoid false positives due to visual occlusion.
To test the scalability of our algorithm, we used two additional syntactically-rich \scenic{} programs, see Fig.~\ref{fig:scalability1} and ~\ref{fig:scalability3}. 
We increased the scenario complexity by maintaining the same program structure while scaling the number of agents. For evaluation, we iteratively fixed the number of agents in a scenario and generated 10 labels using the \scenic{} programs. 

We plot the average runtime results for querying these 10 labels with our algorithm for each fixed agent number in Fig.~\ref{fig:scalability1_chart} and ~\ref{fig:scalability3_chart}. In each plot, we compare two setups of our algorithm: the left bar for the case where  the object correspondence between the objects in the \scenic{} program and the label are \textit{known}, versus the right bar where the correspondence is \textit{unknown}. For the unknown case, we randomly shuffled the ordering of the objects in the labels. 

We observe a consistent behavior from these two plots. When the correspondence is unknown, the runtime increases exponentially in the number of agents due to the combinatorial object correspondence matching process in our algorithm (refer to Sec.~\ref{sec:modular_approach}). On the other hand, when the object correspondence is known, we consistently observed that the runtime of the query algorithm increases approximately linearly in the number of agents, and scales to a sizeable number.

% , the runtime increases exponentially in the number of agents in Fig.~\ref{fig:scalability1_chart} as expected. On the contrary, in Fig.~\ref{fig:scalability3_chart}, the runtime linearly increases with the number of agents even though the correspondence is unknown. In fact, the runtime is approximately the same as the case when the correspondence is known. This is because all the pedestrians in the scenario (shown in Fig.~\ref{fig:scalability3}) are sampled from the same region defined by \texttt{offset by Range(-5,5) @ Range(0,200)} with the same heading constraint as in line 4. Hence, in this unique case, any object correspondence is valid. This dichotomy in the runtime behavior illustrates that, depending on the structure of a \scenic{} program, the algorithm's runtime may not always grow exponentially in the number of the agents when querying the real labelled data whose object correspondence is unknown. On a separate note, in both plots, we observe that the runtime for correspondence unknown case times out at 10 seconds. The reason for this timeout is due to explosion in creating a permutation of all correspondences, which for 22 agent case, there are $22! > 10^{21}$.

%% file: sections/related.tex
\section{Related Work}\label{sec:relatedwork}
The most relevant study~\cite{itsc} to our work was reported where failure scenarios of an autopilot in simulation were validated by \textit{physically} reconstructing them at a track testing facility. However, this manual validation approach is labor-intensive and not scalable. On the contrary, our approach aims to automate this validation in a data-driven manner. 

Domain adaption~\cite{domain_adaptation} aims to reduce the sim-to-real gap from \textit{training} perspective, whose objective is primarily to enhance the performance of a perception model for a particular task (e.g. segmentation, detection, localization) on the \textit{real} sensor data by only training on the simulated sensor data~\cite{sim-to-real1,sim-to-real2,sim-to-real3,sim-to-real4}. Generative adversarial networks (GANs)~\cite{gan} have been a key technique employed to \textit{adapt}, or convert, synthetic data to more realistic data of various types such as RGB images, 3D LiDAR point cloud, etc. These data are used for training. On the contrary, our paper aims to reduce the gap from \textit{testing} perspective, which investigates given an already \textit{pre-trained} model, how does it behave differently in the same scenario across two different domains: simulated and real data.  

Visual question answering (VQA)~\cite{DBLP:journals/ijcv/GoyalKASBP19,DBLP:conf/eccv/XuS16} considers answering questions about static images phrased in natural language. The VQA area combines approaches common in captioning with a large natural language processing component: part of the challenge is to understand the question. Much like VQA, we decide whether an image matches a given query. However, our queries are expressed using a formal probabilistic programming language and we query the \textit{label}, not the sensor data. This allows us to formulate a well-defined querying problem and develop an algorithm which is guaranteed to be sound: the returned subset of images are exactly those which match the scenario program, if the labels are accurate.

Our approach can be seen as a specialised form of inference in probabilistic programming languages (PPLs). \scenic{} allows making probabilistic assertions of propositional statements, e.g., `the car is within the visible region'. Such declarative \emph{hard constraints} make scenario modeling much easier and more intuitive. Some PPLs, e.g.~\textsc{Angluin}, actively prevent specifying hard constraints to prevent programmers from `accidentally' posing NP-hard questions~\cite{DBLP:conf/aistats/WoodMM14}.
PPLs such as \textsc{Pyro}~\cite{DBLP:journals/jmlr/BinghamCJOPKSSH19} and \textsc{Edward}~\cite{DBLP:conf/iclr/TranHSB0B17} use Bayesian inference schemes that require tracking derivatives~\cite{DBLP:journals/jmlr/HoffmanG14,DBLP:journals/corr/abs-1301-1299,DBLP:conf/nips/KucukelbirRGB15,DBLP:journals/jmlr/KucukelbirTRGB17}. Some PPLs allow hard constraints and non-continuity, but either  have (significant) restrictions or limited efficiency~\cite{DBLP:conf/aaai/NoriHRS14,DBLP:journals/corr/abs-1804-03523,DBLP:journals/corr/abs-2005-09089}, putting trigonometry and continuous domains out of reach. 
% Moreover, the typical inference task is to  compute posterior distributions relative to a prior, whereas we are primarily interested in filtering using a Boolean membership query. Finally, we note that while sampling-based approaches may do well in answering membership queries positively, they are not well suited for providing negative answers.

%% file: sections/conclusion.tex
\begin{figure}
    \centering
    \includegraphics[width=\columnwidth]{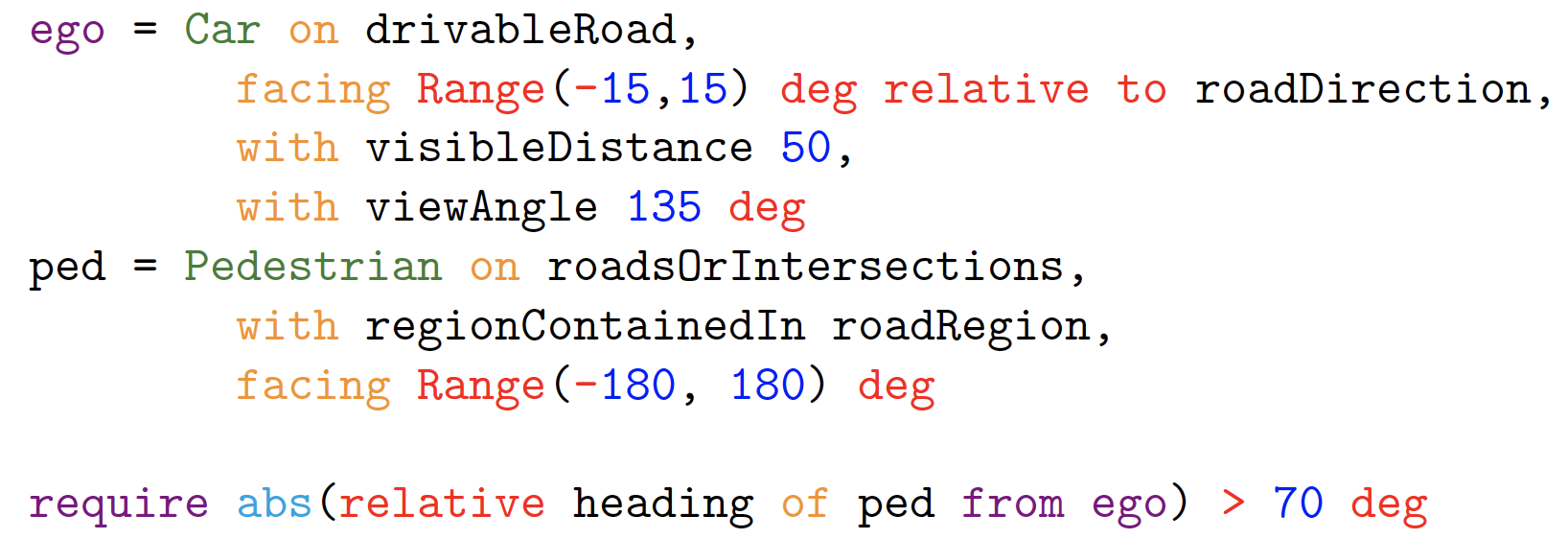}
    \caption{Scenario\# 1 in the Efficacy Experiment}
    \label{fig:efficacy1}
\end{figure}
\begin{figure}
    \centering
    \includegraphics[width=\columnwidth]{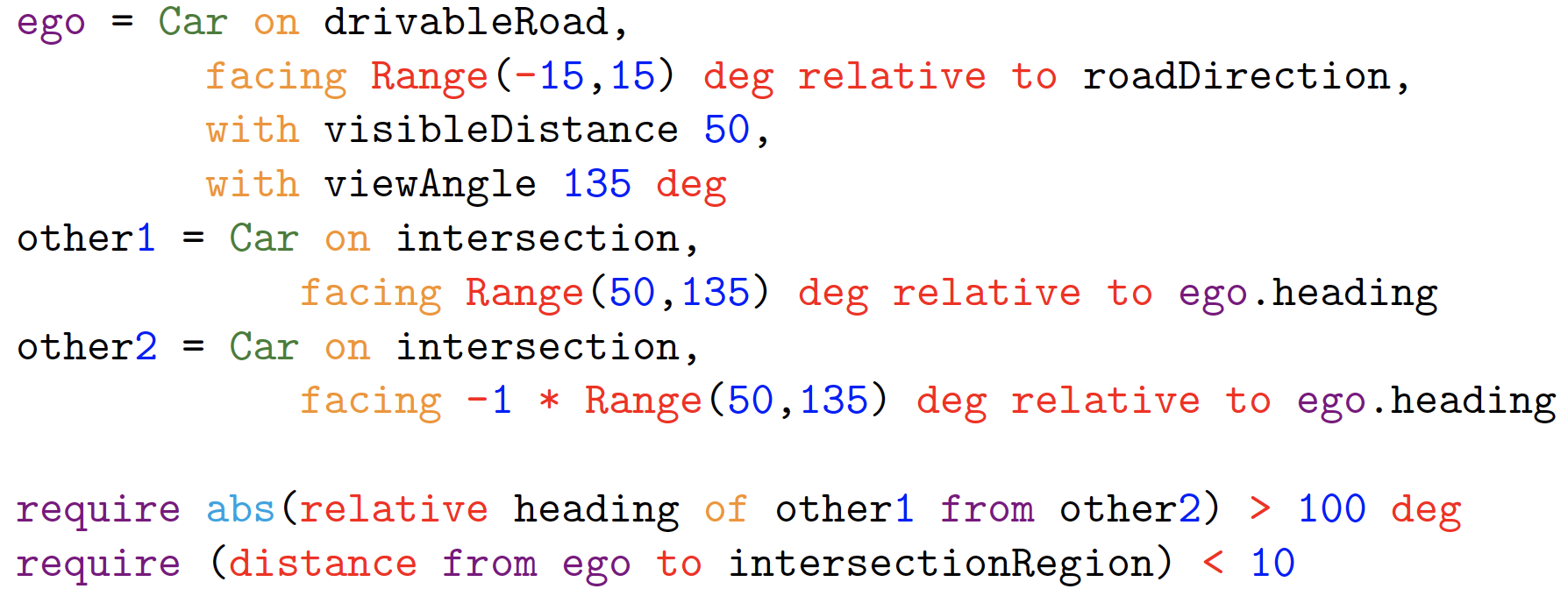}
    \caption{Scenario\# 2 in the Efficacy Experiment}
    \label{fig:efficacy2}
\end{figure}
\begin{figure}
    \centering
    \includegraphics[width=\columnwidth]{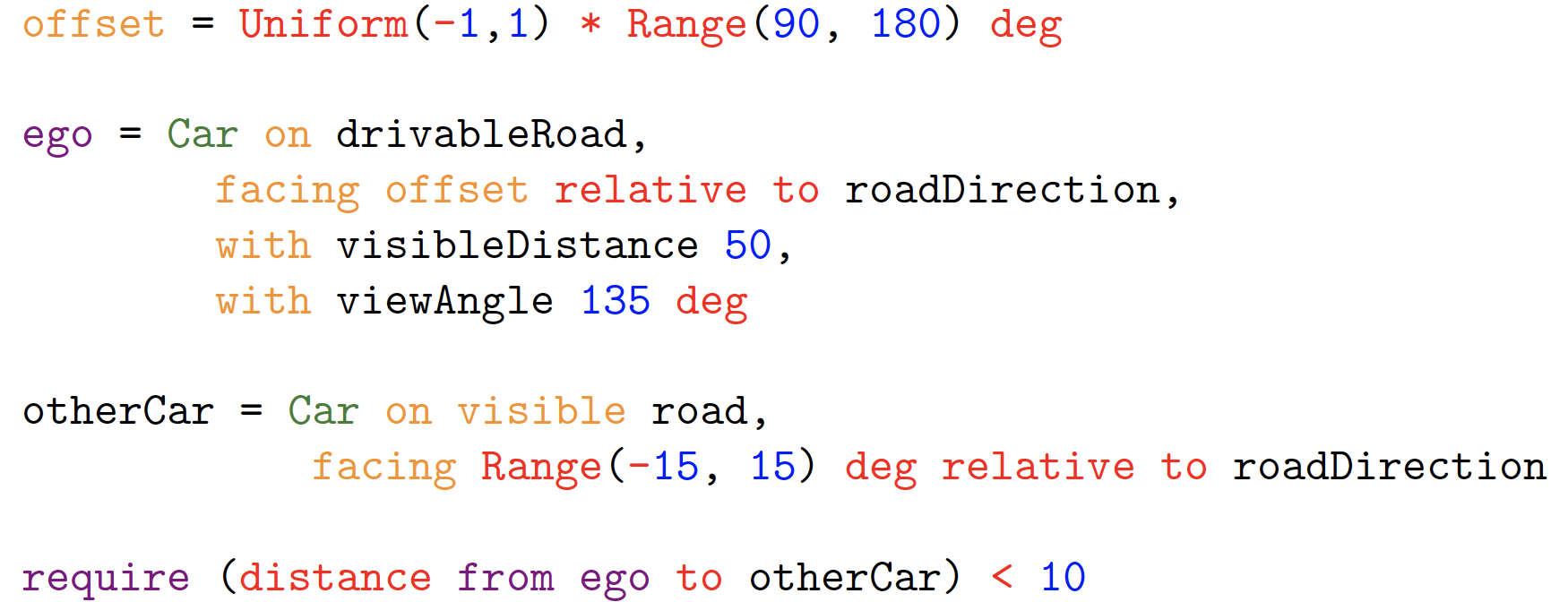}
    \caption{Scenario\# 3 in the Efficacy Experiment}
    \label{fig:efficacy3}
\end{figure}
\begin{figure}
    \centering
    \includegraphics[width=\columnwidth]{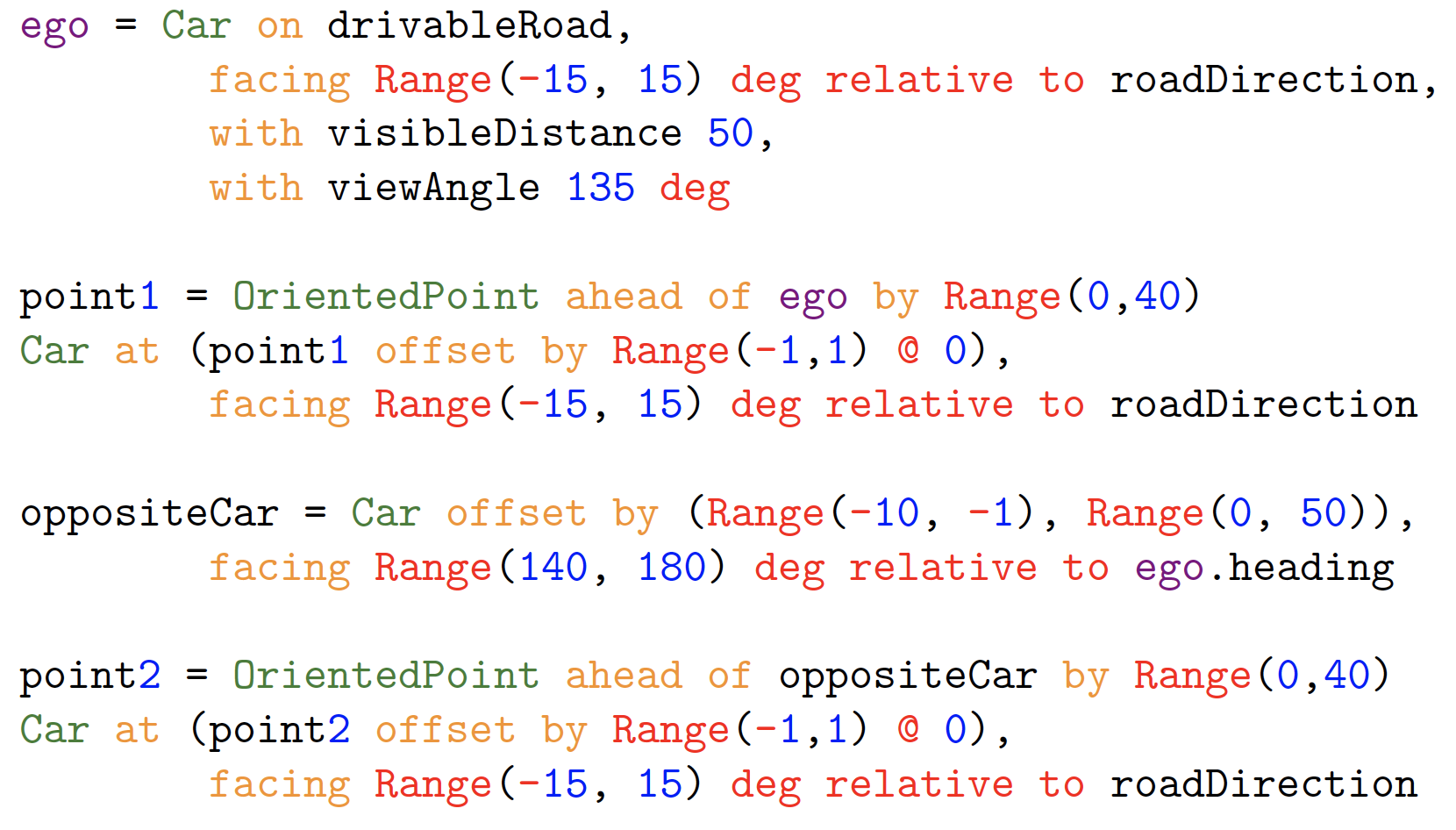}
    \caption{Scenario\# 4 in the Efficacy Experiment}
    \label{fig:efficacy4}
\end{figure}
\begin{figure}
    \centering
    \includegraphics[width=\columnwidth]{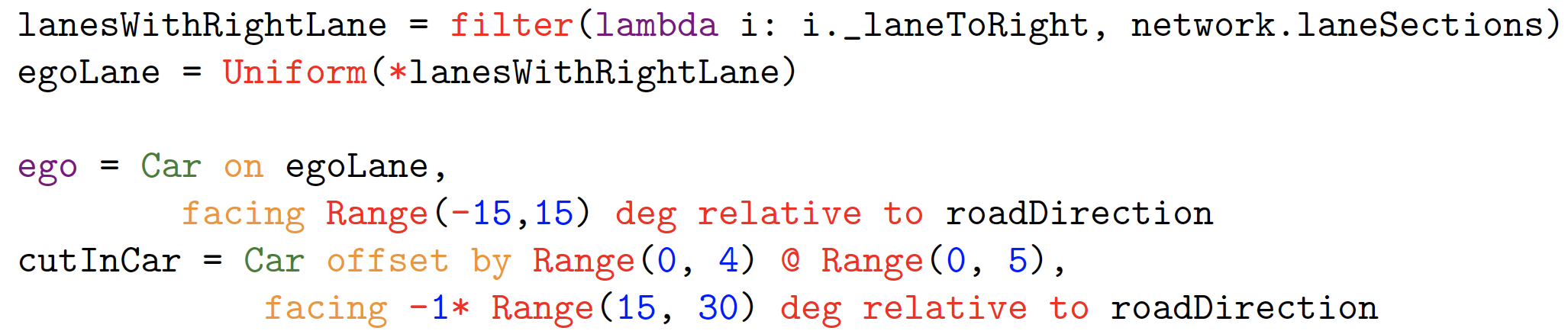}
    \caption{Scenario\# 5 in the Efficacy Experiment}
    \label{fig:efficacy5}
\end{figure}

\section{Conclusion}\label{sec:concl}
We proposed an algorithm to query a labeled dataset using a scenario program encoded in the \scenic{} language.
This algorithm can be used to shrink the gap between simulation-based and real-world testing by identifying counterparts of simulated scenarios in real data, which can then be used to validate the fidelity of the simulations.
More broadly, our algorithm enables a principled way to explore and understand the range of scenarios present in a dataset by expressing scenarios of interest in a formal language.
In future work, we plan to explore using program analysis and other techniques to alleviate the combinatorial explosion when the object correspondence is unknown, and to generalize our algorithm to support \emph{dynamic} scenarios as well as \emph{probabilistic} queries that take the likelihood of labels into account.

%% file: sections/appendix.tex
\newcommand{\mkvec}[2]{\left\langle#1,#2\right\rangle}
\newcommand{\angleto}[2]{\arctan\,(#1-#2)}
\newcommand{\rot}[2]{\textit{rotate}\,(#1,#2)}
\newcommand{\disc}[4]{\textit{Disc}\,(#1,#2,#3,#4)}
\newcommand{\sector}[6]{\textit{Sector}\,(#1,#2,#3,#4,#5,#6)}
\newcommand{\bounding}[1]{\textit{boundingBox}\,(#1)}
\newcommand{\pointin}[1]{\textit{uniformPointIn}\,(#1)}
\newcommand{\visible}[3]{\textit{visibleRegion}\,(#1,#2,#3)}
\newcommand{\orientation}[1]{\textit{orientation}\,(#1)}
\newcommand{\mkop}[2]{\texttt{OrientedPoint}\,(#1,#2)}
\newcommand{\opth}[2]{\mkop{#1}{#2}}
\newcommand{\euler}[3]{\textit{forwardEuler}\,(#1,#2,#3)}
\newcommand{\offsetLocal}[2]{\textit{offsetLocal}\,(#1,#2)}
\newcommand{\sem}[1]{\llbracket#1\rrbracket}
\newcommand{\tm}[1]{\text{#1}}
\newcommand{\leftOfLine}[4]{\textit{leftLine}\,(#1,#2,#3,#4)}
\newcommand{\rightOfLine}[4]{\textit{rightLine}\,(#1,#2,#3,#4)}
\newcommand{\lineSmt}[4]{\textit{lineSeg}\,(#1,#2,#3,#4)}
\newcommand{\xyRange}[2]{\textit{xyRangeOfLine}\,(#1,#2)}
\newcommand{\maximum}[2]{\textit{max}\,(#1,#2)}
\newcommand{\minimum}[2]{\textit{min}\,(#1,#2)}
\newcommand{\maxX}[2]{\textit{maxX}\,(#1,#2)}
\newcommand{\minX}[2]{\textit{minX}\,(#1,#2)}
\newcommand{\maxY}[2]{\textit{maxY}\,(#1,#2)}
\newcommand{\minY}[2]{\textit{minY}\,(#1,#2)}
\newcommand{\slope}[2]{\textit{slope}\,(#1,#2)}
\newcommand{\offset}[2]{\textit{offset}\,(#1,#2)}
\newcommand{\rangeX}[2]{\textit{rangeX}\,(#1,#2)}
\newcommand{\rangeY}[2]{\textit{rangeY}\,(#1,#2)}
\newcommand{\triangleRegion}[5]{\textit{tri}\,(#1,#2,#3,#4,#5)}

\section{SMT Encoding of the \scenic{} Fragment}
\subsection{Supported \scenic{} Fragment}\label{appendix:fragment}
We support a large fragment of \scenic{} but exclude some language features. In particular, our \scenic{} fragment includes 26 different position and heading specifiers as well as scalar, Boolean, heading, vector, region, and \texttt{OrientedPoint} operators, which together provides expressive language to flexibly model wide variety of scenarios.
That is, the \scenic{} fragment we use for our SMT encoding, which enables our proposed querying, include all \scenic{} syntax except for the following cases: (1) the operators \texttt{following F for S} and \texttt{following F from V for S}, (2) any require statement's reference to any variables (i.e. semantic features) that are not present in the label, 
% the semantic features referenced in the \texttt{require} statements must be present in the given label and jointly dependent variables (defined in Sec.~\ref{sec:dependency_analysis}) occurring within \texttt{require} statements, 
and (3) imports of external Python libraries.

\subsection{SMT Encoding}\label{appendix:encoding}
To succinctly present our encoding, we first introduce some notations. First, we denote any \scenic{} semantics with a double bracket, $\sem{\cdot}$ and denote SMT encoding of any expression as $E(\cdot)$. For example, to access a position of a object \textit{O} as we would in \scenic{}, we denote $\sem{O.position}$ which is equivalent to the object's x, y coordinate, $\mkvec{O_x}{O_y}$. \scenic{} employs an ego-centric syntax, meaning it requires that \texttt{ego} be defined and its syntax, by default, assumes ego as a reference object if not otherwise specified. We will use \textit{O} to represent a \scenic{} object, heading as \textit{H}, vector (i.e. position) as \textit{V}. $\top$ denotes True. Basic notations are shown in Fig.~\ref{figure:notations}.

A \scenic{} expression can be categorized into three types: (1) built-in functions, (2) predicates, and (3) distributions. For each expression, we create a new SMT variable, encode any constraints on its value implied by the \scenic{} semantics, and gather these variables and encoded constraints throughout each incremental encoding process. For example, to encode the expression \texttt{Range(2,5)}, we create a new SMT variable \textit{range1} and add the formula $2\leq \textit{range1}\leq 5$ to our set of constraints.
Our SMT encoding for \scenic{}'s built-in distributions are shown in Fig.~\ref{figure:distributions}.

Second, all the specifiers and operators in our \scenic{} fragment are built-in functions in \scenic{}. Therefore, they can be abstractly represented in the following form: $f(a_1,...,a_k)$ where the function, $f$, represents the \scenic{} specifiers and operators, and $a_1,...,a_k$ are input arguments, where each input argument can be \scenic{} expression. We first encode the input arguments to SMT formula and then encode the function according to the semantics of the specifiers and operators, as defined in the Appendix C.2 through C.5 in the \scenic{} paper~\cite{scenic-journal}. Hence, the SMT encoding of the abstract form: \\
$E(f(a_1,...,a_k)) = \sem{f}(E(a_1),...,E(a_k))$. It is possible that these arguments may themselves consist of specifiers and operators with additional input arguments, creating an tree of syntax with its leaf nodes being the input arguments. In such a case, we traverse down to the leaf nodes and recursively encode the tree by traversing toward the root node. 

For example, the SMT formula for the position of \texttt{otherCar} in the \scenic{} program in Fig.~\ref{fig:scenic_program} is: \\
$\sem{\text{ahead of}}(E(ego.position),E(ego.heading),E(Range(4,10)))$. Here,\\ $E(ego.position)$ is a \scenic{} expression which represents $\sem{\texttt{On}}(\texttt{road})$. In this case, we would recursively traverse down to the leaf node(s) of each input argument. For the syntax tree created by ego position, the leaf node is \texttt{road} region. We encode this region, then encode the predicate \texttt{On} to encode $E(ego.position)$. Likewise, we encode the other input arguments and then lastly $\sem{\texttt{ahead of}}$.

Finally, the SMT encoding for predicates in \scenic{} and all the regions in the map are shown in Fig.~\ref{figure:regions}. All the regions in the map, such as road and lane regions represented as polygons which are triangulated, meaning that each region are decomposed into a set of triangles using pypoly2tri~\cite{mapbox} Python package. And, these triangle regions are encoded as SMT formula.

% Given these encoding schemes over \scenic{} specifiers and operators, a full \scenic{} program is encoded into SMT formula in the following way. As mentioned in Sec.~\ref{sec:method}, a \scenic{} program is compiled into an expression forest with its root nodes representing semantic features. In the order of dependency, each (or a set of, if jointly dependent) semantic feature expression tree(s) is incrementally encoded into a SMT formula and checked for satisfaction. This SMT encoding is recursively done from the leaf nodes of the semantic feature tree toward the root. Refer to Fig.~\ref{fig:expression_tree} and Sec.~\ref{sec:overview} for a concrete example. In this encoding scheme, we can also encode any \scenic{} program with nested specifiers and/or operators, for example, as shown in Fig.~\ref{fig:shared_variable}.
 
\begin{figure}
\begin{align*}
\mkvec{x}{y} &= \text{point with the given XY coordinates} \\
E(V) &= \mkvec{E(V_x)}{E(V_y)} \\
E(V_1)\pm E(V_2) &= \mkvec{E(V_{1,x}) \pm E(V_{2,x})}{E(V_{1,y}) \pm E(V_{2,y})} \\
k*(E(V_1)\pm E(V_2)) &= \mkvec{k*(E(V_{1,x})\pm E(V_{2,x}))}{k*(E(V_{1,y})\pm E(V_{2,y}))},\\ & \text{where } k \in \mathbb{R}\\
\rot{\mkvec{x}{y}}{\theta} &= \langle E(x) \cos (E(\theta)) - E(y) \sin (E(\theta)),\\
& E(x) \sin (E(\theta)) + E(y) \cos (E(\theta)) \rangle\\
\offsetLocal{O}{v} &= \sem{\tm{O.position}} + \rot{v}{\sem{\tm{O.heading}}}\\
\textit{OP}(V,\theta) &= \text{\texttt{OrientedPoint} with position V and heading } \theta
\end{align*}
\caption{Notation used to define the SMT Encoding of \scenic{} syntax}
\label{figure:notations}
\end{figure}

\begin{figure}
\begin{align*}
Range(l,u) &= E(l) \leq \text{z} \leq E(u) \\
Normal(m,s) & = \top \\
Option(a_1, a_2, ..., a_n) &= (z==E(a_1))\vee(z==E(a_2))\vee...\vee(z==E(a_n))
\end{align*}
\caption{The SMT encoding of distributions in \scenic{} where z is the new SMT variable constructed for each distribution}
\label{figure:distributions}
\end{figure}

\begin{figure}
\begin{align*}
\maxX{V_1}{V_2} &= \maximum{E(V_{1,x})}{E(V_{2,x})};\\ &\maxY{V_1}{V_2} \text{ defined likewise}\\
\minX{V_1}{V_2} &= \minimum{E(V_{1,x})}{E(V_{2,x})};\\  &\minY{V_1}{V_2} \text{ defined likewise}\\
\rangeX{V_1}{V_2} &= [\minX{V_1}{V_2},\maxX{V_1}{V_2}]\\
\rangeY{V_1}{V_2} &= [\minY{V_1}{V_2},\maxY{V_1}{V_2}]\\
\slope{V_1}{V_2} &= (E(V_{2,y})-E(V_{1,y})) / (E(V_{2,x})-E(V_{1,x}))\\
\offset{V_1}{V_2} &= E(V_{1,y})-\slope{E(V_1)}{E(V_2)}*E(V_{1,x})\\
\lineSmt{V_1}{V_2}{x}{y} &= \text{a point} \mkvec{x}{y} \text{ is on the line segment} \\
& (y == \slope{V_1}{V_2}*x + \offset{V_1}{V_2}),\\
& \text{if x}\in \rangeX{V_1}{V_2}, \text{y}\in \rangeY{V_1}{V_2}\\
\leftOfLine{V_{1}}{V_{2}}{x}{y} &= \text{a point} \mkvec{x}{y} \text{ is on the left region of the line}\\
& \text{ whose direction is } V_1 \text{ to } V_2 \\
&= D_x * T_y - D_y * T_x > 0, \\
& \text{ where } D=E(V_2)-E(V_1), T=\mkvec{x}{y}-E(V_1)\\
\disc{c}{r}{x}{y} &= \text{a point} \mkvec{x}{y} \text{ in a circle centered at c}\\
&= ((x-E(c_x))^2+(y-E(c_y))^2 \leq E(r)^2)\\
\sector{c}{r}{h}{a}{x}{y} &= \text{a point} \mkvec{x}{y} \text{on a sector of } \disc{c}{r}{x}{y}\\ & \text{ heading h with angle a}\\
&= \disc{c}{r}{x}{y}\land\rightOfLine{c}{V_1}{x}{y}\\
& \land\leftOfLine{c}{V_2}{x}{y},\\
& \text{where } V_1=\offsetLocal{\textit{OP}(c,h-a/2)}{\mkvec{0}{r}}, \\
& V_2=\offsetLocal{\textit{OP}(c,h+a/2)}{\mkvec{0}{r}}\\
% \bounding{O} &= \text{set of points in the bounding box of object } O \\
\visible{X}{x}{y} &= \begin{cases}
\sector{\sem{\tm{X.position}}}{\sem{\tm{X.viewDistance}}}{\\\quad\quad\quad\sem{\tm{X.heading}}}{\sem{\tm{X.viewAngle}}}{x}{y},\\
\text{if X is } \texttt{OrientedPoint} \\
\disc{\sem{\tm{X.position}}}{\sem{\tm{X.viewDistance}}}{x}{y}, \\
\text{if X is } \texttt{Point}
\end{cases} \\
\triangleRegion{V_0}{V_1}{V_2}{x}{y} &= \text{a point} \mkvec{x}{y} \text{in a triangle region consisting}\\
& \text{of three points}\\
&=(\mkvec{x}{y}==E(V_1)+(E(V_1)-E(V_2))*s\\
& +(E(V_2)-E(V_0))*t),\\
& \text{where } \exists s,\exists t, 0\leq s\leq 1, 0\leq t\leq 1, s+t\leq 1, \\
& \text{using barycentric coordinate system~\cite{barycentric}} \\
\sem{On}(\text{region}, \mkvec{x}{y}) &= \bigvee\limits_{i=1}^{n}  \triangleRegion{V_0}{V_1}{V_2}{x}{y},\\
& \text{for all triangles in a region of a map}, \\
& \text{where $V_0,V_1,V_2$ compose each different triangle}
\end{align*}
\caption{SMT encoding of regions where x, y are the new SMT variable constructed for a region}
\label{figure:regions}
\end{figure}

\section{Proof Sketch of Theorem 1}\label{appendix:proof}
Due to the page limit on the Appendix, we provide a sketch of our proof. \\ 
\textbf{Proposition 1: The monolithic encoding algorithm returns \texttt{sat} if and only if the label \textit{matches} the program for a given object correspondence}\\
pf) Based on this assumption, we prove by contrapositive. Suppose the algorithm returns \texttt{unsat}. Then, either the monolithic encoding of the program, excluding any require statement, or any require statement conditioning on the observed feature values is \textit{not} satisfied. We assume that the monolithic encoding (based on our SMT encoding in Appendix~\ref{appendix:encoding} is correctly encodes ) is correct and the require statement encoding by the \scenic{} compiler is correct, then there exists a feature in the label that is not satisfying its constraints. 
Suppose 

Therefore, the SMT encoding of the \scenic{} program excluding the require statement is correct. This SMT formula can be checked with an SMT solver. If satisfied, all the semantic features can be conditioned to the observed values in the label with a given object correspondence and any require statements are checked for satisfaction. If require statements are also satisfied, the \scenic{} program matches the label. 

\textbf{Proposition 2: Our incremental SMT encoding of (jointly) dependent semantic features in the \scenic{} program is equivalent to monolithic encoding of the program, excluding the require statements for a given correspondence} \\
pf) Suppose there are n semantic features, $s_1, s_2, ..., s_n$, in the label and the program. We denote $\phi_{1}, \phi_{2}, ..., \phi_{n}$ as the corresponding SMT encoding of the semantic features and require statements, respectively. For brevity, assume that these SMT encoding of the semantic features already compares the semantic features constraints to corresponding semantic feature values observed in the label. Then, the monolithic encoding of the program is equivalent to $\phi_{1}\wedge \phi_{2} \wedge \cdot\cdot\cdot \wedge \phi_{n}$. 

For simplicity, let's first consider a \scenic{} program with only dependent semantic features, and contains no \textit{jointly} dependent features. Suppose our dependency analysis step returned $s_1, s_2, ..., s_n$. This dependency, simply put, represents a containment relation (refer to Sec.~\ref{sec:dependency_analysis}). For example, if $s_2$ is dependent on $s_1$, this means that $s_2$'s expression tree contains that of $s_1$. Then, our incremental encoding is equivalent to \\
$\phi_{1}\wedge \phi_{2}\vert \phi_{1}\wedge \phi_{3}\vert(\phi_{2},\phi_{1}), \cdot\cdot\cdot \phi_{n}\vert(\phi_{n-1},\phi_{n-2},\cdot\cdot\cdot,\phi_{1})$ \\
where the vertical bar, $\vert$, represents conditioning of the semantic features to the corresponding observed values in the label (as discussed in Sec.~\ref{sec:overview}). Basically, the SMT formula, $\phi_{i}\vert (\phi_{i-1},...,\phi_{1})$ encodes $s_i$ while conditioning (i.e. substituting) the $s_1, ..., s_{i-1}$'s expression trees which are contained in $s_i$'s expression tree with their corresponding observed values in the label. Due to this containment relation, the following rule naturally follows: 
$\phi_{1}\wedge \phi_{2} = \phi_{1}\wedge \phi_{2}\vert \phi_{1}$. Applying this rule iteratively, we derive the following:\\ 
$\phi_{1}\wedge \phi_{2} \wedge \cdot\cdot\cdot \wedge \phi_{n}$ \\
$=\phi_{1}\wedge \phi_{2}\vert \phi_{1}\wedge \phi_{3}\vert(\phi_{2},\phi_{1}), \cdot\cdot\cdot \phi_{n}\vert(\phi_{n-1},\phi_{n-2},\cdot\cdot\cdot,\phi_{1})$ \\
Therefore, in this case, the incremental SMT encoding of dependent semantic features is equivalent to the monolithic encoding of the program. 

This result extends to case when semantic features are also include jointly dependent relations. Suppose all m semantic features are jointly dependent. For the convenience of notation, suppose $s_i, s_{i+1}, ..., s_{i+m-1}$ where $m\leq n$ and $i\geq 1$ are jointly dependent, although their numbering does not have to be consecutive. Then, the incremental SMT encoding of the program is equivalent to \\ \\
$\phi_{1}\wedge \phi_{2}\vert \phi_{1}\wedge\cdot\cdot\cdot\wedge (\phi_{i}\wedge\phi_{i+1}\wedge\cdot\cdot\cdot\wedge\phi_{i+m-1})\vert(\phi_{i-1},\cdot\cdot\cdot,\phi_{1})\\ 
\wedge\phi_{i+m}\vert(\phi_{i+m-1},\cdot\cdot\cdot,\phi_{1})\wedge\cdot\cdot\cdot\wedge\phi_{n}\vert(\phi_{n-1},\phi_{n-2},\cdot\cdot\cdot,\phi_{1})$\\ 
$= \phi_{1}\wedge\phi_{2}\wedge\cdot\cdot\cdot\wedge\phi_{i-1}\wedge(\phi_{i}\wedge\phi_{i+1}\wedge\cdot\cdot\cdot\wedge\phi_{i+m-1})\vert(\phi_{i-1},\cdot\cdot\cdot,\phi_{1})\\ 
\wedge\phi_{i+m}\vert(\phi_{i+m-1},\cdot\cdot\cdot,\phi_{1})\wedge\cdot\cdot\cdot\wedge\phi_{n}\vert(\phi_{n-1},\phi_{n-2},\cdot\cdot\cdot,\phi_{1})$\\ 
$=\phi_{1}\wedge\phi_{2}\wedge\cdot\cdot\cdot\wedge\phi_{i-m+1}\wedge\phi_{i+m}\vert(\phi_{i+m-1},\cdot\cdot\cdot,\phi_{1})\wedge\cdot\cdot\cdot\wedge\phi_{n}\vert(\phi_{n-1},\phi_{n-2},\cdot\cdot\cdot,\phi_{1})$\\
$=\phi_{1}\wedge \phi_{2} \wedge \cdot\cdot\cdot \wedge \phi_{n}$ \\

by applying the iteratively applying the rule, $\phi_{1}\wedge \phi_{2} = \phi_{1}\wedge \phi_{2}\vert \phi_{1}$. Then, this results trivially extends to case when there exists more than one jointly dependent set of semantic features. Therefore, the incremental SMT encoding is equivalent to the monolithic encoding of the \scenic{} program, excluding the require statements. \\ \\

\textbf{Proof Sketch of Theorem 1}: \\
By Proposition 2, our algorithm, which incrementally encodes (jointly) dependent semantic features in the order of dependency, equivalently encodes the the \scenic{} program, excluding the require statements, as the monolithic encoding algorithm. Then, by the same reasoning in the proof of Proposition 1, our algorithm is returns \texttt{sat} if and only if the label \textit{matches} the program.